\documentclass{article}

\PassOptionsToPackage{numbers, compress}{natbib}

\usepackage[final]{neurips_2022}

\usepackage[utf8]{inputenc} %
\usepackage[T1]{fontenc}    %
\usepackage{hyperref}       %
\usepackage{url}            %
\usepackage{booktabs}       %
\usepackage{amsfonts}       %
\usepackage{nicefrac}       %
\usepackage{microtype}      %
\usepackage{xcolor}         %

\usepackage{times}
\usepackage{mathtools}
\usepackage[capitalise]{cleveref}
\usepackage{color}
\usepackage{booktabs,siunitx}
\usepackage{comment}

\usepackage{graphicx}
\usepackage{xspace}
\usepackage{capt-of}

\usepackage{amsmath}
\usepackage{amssymb}
\usepackage{amsfonts}
\usepackage{amsthm}

\usepackage{algorithm}
\usepackage[noend]{algpseudocode}

\usepackage{multirow}

\newtheorem{lem}{Lemma}[section]
\newtheorem{thm}{Theorem}[section]

\newtheorem{prop}{Proposition}[section]

\newtheorem{definition}{Definition}[section]
\newtheorem{claim}[lem]{Claim}

\Crefname{thm}{Theorem}{Theorems}
\Crefname{cor}{Corollary}{Corollaries}
\Crefname{lem}{Lemma}{Lemmas}
\Crefname{prop}{Proposition}{Propositions}
\Crefname{assumption}{Assumption}{Assumptions}
\Crefname{definition}{Definition}{Definitions}
\Crefname{claim}{Claim}{Claims}
\Crefname{table}{Table}{Tables}

\newcommand{\HonakerOnline}{\texttt{Honaker Online}\xspace}
\newcommand{\HonakerFull}{\texttt{Honaker Full}\xspace}
\newcommand{\OptPrefixSum}{\texttt{Opt Prefix Sum}\xspace}
\newcommand{\OptMomentum}{\texttt{Optimal M = B C}\xspace}

\definecolor{darkgreen}{rgb}{0,0.4,0.0}
\definecolor{darkblue}{rgb}{0,0.0,0.4}
\definecolor{darkred}{rgb}{0.4,0,0.0}

\newcommand{\bfA}{\ensuremath{\mathbf{A}}}
\newcommand{\bfB}{\ensuremath{\mathbf{B}}}
\newcommand{\bfC}{\ensuremath{\mathbf{C}}}
\newcommand{\bfD}{\ensuremath{\mathbf{D}}}

\newcommand{\bfG}{\ensuremath{\mathbf{G}}}
\newcommand{\bfH}{\ensuremath{\mathbf{H}}}
\newcommand{\bfI}{\ensuremath{\mathbf{I}}}

\newcommand{\bfL}{\ensuremath{\mathbf{L}}}
\newcommand{\bfM}{\ensuremath{\mathbf{M}}}

\newcommand{\bfQ}{\ensuremath{\mathbf{Q}}}
\newcommand{\bfR}{\ensuremath{\mathbf{R}}}
\newcommand{\bfS}{\ensuremath{\mathbf{S}}}
\newcommand{\bfT}{\ensuremath{\mathbf{T}}}
\newcommand{\bfU}{\ensuremath{\mathbf{U}}}
\newcommand{\bfV}{\ensuremath{\mathbf{V}}}
\newcommand{\bfW}{\ensuremath{\mathbf{W}}}
\newcommand{\bfX}{\ensuremath{\mathbf{X}}}

\newcommand{\bfZ}{\ensuremath{\mathbf{Z}}}

\newcommand{\bfa}{\ensuremath{\mathbf{a}}}

\newcommand{\bfc}{\ensuremath{\mathbf{c}}}
\newcommand{\bfd}{\ensuremath{\mathbf{d}}}

\newcommand{\bfg}{\ensuremath{\mathbf{g}}}
\newcommand{\bfh}{\ensuremath{\mathbf{h}}}

\newcommand{\bfm}{\ensuremath{\mathbf{m}}}

\newcommand{\bfs}{\ensuremath{\mathbf{s}}}

\newcommand{\bfv}{\ensuremath{\mathbf{v}}}
\newcommand{\bfw}{\ensuremath{\mathbf{w}}}
\newcommand{\bfx}{\ensuremath{\mathbf{x}}}
\newcommand{\bfy}{\ensuremath{\mathbf{y}}}

\newcommand{\calA}{\ensuremath{\mathcal{A}}}

\newcommand{\calD}{\ensuremath{\mathcal{D}}}

\newcommand{\calL}{\ensuremath{\mathcal{L}}}
\newcommand{\calM}{\ensuremath{\mathcal{M}}}
\newcommand{\calN}{\ensuremath{\mathcal{N}}}
\newcommand{\calO}{\ensuremath{\mathcal{O}}}

\newcommand{\calS}{\ensuremath{\mathcal{S}}}
\newcommand{\calT}{\ensuremath{\mathcal{T}}}

\newcommand{\R}{\mathbb{R}}

\newcommand{\C}{\mathbb{C}}
\newcommand{\E}{\mathbb{E}}

\newcommand{\ip}[2]{\langle #1, #2\rangle}

\newcommand{\dimension}{\ensuremath{n}}
\newcommand{\dimensions}{\ensuremath{n}}

\newcommand{\dimxdim}{\ensuremath{{\dimension\!\times\!\dimension}}}

\DeclareMathOperator{\diag}{diag}
\DeclareMathOperator{\diagpart}{diagpart}

\newcommand{\xopt}{\ensuremath{\bfX^\star}} %
\newcommand{\thetaopt}{\ensuremath{\theta^\star}} %
\newcommand{\obs}{\bfg}  %
\newcommand{\obsp}{\bfh}  %
\newcommand{\obsM}{\bfG} %
\newcommand{\obsMp}{\bfH}  %
\newcommand{\noiseM}{\bfZ}

\newcommand{\mdim}{d}  %
\newcommand{\obsMti}{\obsM\idx{1:i}{:}}
\newcommand{\datadim}{{\dimension \times \mdim}}

\newcommand{\example}{\chi}
\newcommand{\Ctree}{\bfC_{\mathcal{T}}}
\newcommand{\Bhs}{\bfB_{\text{hs}}}
\newcommand{\Bhf}{\bfB_{\text{hf}}}
\newcommand{\ML}{\bfM^{(\eta)}}
\newcommand{\MR}{\bfM^{(\beta)}}

\DeclareMathOperator*{\argmin}{arg\,min}

\newcommand{\tr}{\ensuremath{\mathrm{tr}}}

\newcommand{\clipnorm}{\zeta}
\renewcommand{\epsilon}{\varepsilon}

\newcommand{\inv}{^{-1}}
\newcommand{\tp}{^\top}

\newcommand{\bftheta}{\ensuremath{\boldsymbol\theta}}
\newcommand{\dualv}{\bfv}
\newcommand{\vopt}{\ensuremath{\bfv^\star}} %
\newcommand{\result}{\bfy}
\newcommand{\assign}{:=}
\newcommand{\plusequal}{\mathrel{+}=}
\newcommand{\myindent}{\hspace{2em}}
\newcommand{\Xofv}{\mathcal{X}}

\newcommand{\mypar}[1]{\paragraph{#1}}

\newcommand{\ltwo}[1]{\left\|#1\right\|_2}

\newcommand{\idx}[2]{_{[#1, #2]}}

\newcommand{\side}{\text{\sf side}}
\newcommand{\adv}{\mathcal{A}}
\newcommand{\mech}{\mathcal{M}}

\newcommand{\ex}[2]{{\ifx&#1& \mathbb{E} \else \underset{#1}{\mathbb{E}} \fi \left(#2\right)}}
\newcommand{\pr}[2]{{\ifx&#1& \mathbb{P} \else \underset{#1}{\mathbb{P}} \fi \left(#2\right)}}
\newcommand{\var}[2]{{\ifx&#1& \mathrm{Var} \else \underset{#1}{\mathrm{Var}} \fi \left(#2\right)}}

\newcommand{\eps}{\varepsilon}
\renewcommand{\epsilon}{\eps}

\newcommand{\zo}{\{0,1\}}
\newcommand{\bit}[1]{{\zo}^{#1}}

\newcommand{\norm}[2]{\left \|{#1}\right \|_{#2}}

\newcommand{\paren}[1]{{\left ( {#1} \right)}}

\newcommand{\id}{\mathbb{I}}

\newcommand{\cN}{\mathcal{N}}

\newcommand{\nul}{\bot}
\newcommand{\power}[2]{\left(#1\right)^{#2}}

\newtheoremstyle{named}{}{}{\itshape}{}{\bfseries}{.}{.5em}{\thmnote{#3 Restated}}
\theoremstyle{named}

\title{Improved Differential Privacy for SGD via Optimal Private Linear Operators on Adaptive Streams}

\author{
  Sergey Denisov \\
  University of Wisconsin-Madison \\
  \texttt{denissov@wisc.edu} \\
  \And
  H. Brendan McMahan \\
  Google Research \\
  \texttt{mcmahan@google.com} \\
  \And
  Keith Rush \\
  Google Research \\
  \texttt{krush@google.com} \\
  \And
  Adam Smith \\
  Boston University \\
  \texttt{ads22@bu.edu} \\
  \And
  Abhradeep Thakurta \\
  Google Research \\
  \texttt{athakurta@google.com} \\
}

\begin{document}

\maketitle

\begin{abstract}
      Motivated by recent applications requiring differential privacy over adaptive streams, we investigate optimal instantiations of the matrix mechanism \citep{Li2015TheMM} in this setting. We prove fundamental theoretical results on the applicability of matrix factorizations to adaptive streams, and provide a parameter-free fixed-point algorithm for computing optimal factorizations. We instantiate this framework with respect to concrete matrices which arise naturally in machine learning, and train user-level differentially private models with the resulting optimal mechanisms, yielding significant improvements in a notable problem in federated learning with user-level differential privacy.
\end{abstract}

\section{Introduction and background}\label{sec:intro}

An important setting for private data analysis is that of streaming inputs and outputs---often dubbed \textit{continual release}. Hiding individual information is especially challenging in such settings since the arrival of one person's data may affect all future outputs of the system. A significant line of work formalizes  \textit{differential privacy} (DP, \citep{10.1007/11681878_14}) under continual release and builds algorithms that meet the resulting definition (e.g.~\citep{CSS11-continual,Dwork-continual,JKT-online,kairouz2021practical,thakurta2013nearly,JainRSS21}). One prominent application of private, continual-release algorithms is to adapt iterative optimization algorithms such as SGD so that their outputs satisfy DP~\citep{thakurta2013nearly,kairouz2021practical,agarwal2017price}.

The problem of privately computing \textit{cumulative sums} plays a key role in both theory and applications. 
Given a set of input vectors (e.g., gradients) $\obs_1,\dots,\obs_\dimension$ with $\obs_i\in\mathbb{R}^\mdim$, the task is to approximate the sequence of  prefix sums 
$(\obs_1, \obs_1 + \obs_2, \dots, \obs_1 + \cdots + \obs_n)$ while satisfying DP. Solutions to this task form the core building block in DP algorithms for online PCA~\citep{dwork2014analyze}, online marginal estimation~\citep{Dwork-continual,CSS11-continual,cormode2019answering}, online top-k selection~\citep{CR21}, and training ML models~\citep{smith2017interaction, kairouz2021practical,SmithThakurta13}, among others.
For example, a common approach to private optimization is to add noise to the gradient estimates in SGD \citep{song2013stochastic,BST14,dpsgd_2016}. \citet{kairouz2021practical} make the observation that the key DP primitive in such contexts is not the independent estimation of individual gradients, but rather the accurate estimation of cumulative sums of gradients. Lowering the error of the DP algorithm's approximation to the cumulative sum translates directly to improved optimization.\footnote{SGD with constant learning rate $\eta$ serves as an intuitive illustration: performing SGD on parameters $\theta$ starting from $\mathbf{0}$, the $t$-th iterate is simply $\theta_t = -\eta \sum_{i=1}^t \obs_i$, where $\obs_i$ is the gradient computed on step $i$. That is, the learned model parameters $\theta_t$ are exactly a scaled version of the cumulative sum of gradients so far. It is the total error in these cumulative sums that matters most, not the error in the private estimates of each individual $\obs_i$. See Theorem 5.1 of \citep{kairouz2021practical} for a formal statement.} 

The structure of continual-release algorithms imposes two major constraints: first, the algorithm must be computable online—that is, we must produce the output at a given time using only prior inputs---and second, privacy analysis must account for \textit{adaptively defined inputs}---that is, the guarantee should hold even against an adversary that selects inputs based on all previous outputs of the system. In learning applications, the privacy analysis must be adaptive even when the stream or raw training examples is fixed in advance, because the points at which we compute gradients depend adaptively on the output of the mechanism so far~\citep{SmithThakurta13}.

In this paper, we revisit the design of continual-release algorithms for cumulative sums and related problems. We give new tools for analyzing privacy in the adaptive setting, new methods to design optimal (within a class) algorithms for summation-style problems, and applications to central problems in private machine learning. Although we focus on learning as the primary application of our algorithmic and analytic tools, our techniques apply to a wide range  of private computations over streaming data such as online monitoring~\cite{tyagi2017stadium,erlingsson2019private}, tracking distributional changes over data streams~\cite{joseph2018local}, and detection of emerging trends~\cite{apple_report}. 

\paragraph{Prior Approaches} Prior approaches to DP approximation of cumulative sums fall broadly into two categories. First, in streaming settings, existing work is generally based on the \textit{binary-tree estimator} \citep{Dwork-continual,CSS11-continual}. This estimator embeds the values to be summed as leaf nodes in a complete binary tree $\calT$, with internal nodes representing the sum of all leaves below them. The mechanism views the \emph{entire tree} as the object to be privately released (ensuring privacy by adding independent noise to each node). An important refinement of \citet{honaker_trick} leverages the multiple independent noisy observations of correlated values to produce lower-variance estimates of the prefix sums. \citet{kairouz2021practical} apply Honaker's online variant, dubbed \HonakerOnline, to the \textit{follow-the-regularized-leader} approach to optimization~\citep{mcmahan2010adaptive,mcmahan2011follow,duchi2011adaptive}. The tree structure of these mechanisms allows for privacy analysis in the adaptive setting, as observed by \cite{SmithThakurta13} and formalized by \cite{JainRSS21}.

The second, more general approach to cumulative sums has previously only been applied in \textit{offline} settings, in which the input is received and outputs are produced as a single batch. The idea is to view cumulative sums as a special case of linear query release (since each output is a pre-specified linear combination of the inputs).  In the offline setting, the tree-based approaches can be viewed as instantiations of this widely-studied \textit{factorization} framework (as in, for example,  \citep{HT10,Li2015TheMM,opt_convex_fact,hdmm,edmonds_nikolov_ullman}).
To introduce the general matrix factorization approach, consider the task of computing a private estimate of a linear mapping $\obsM \mapsto \bfA\obsM$ defined by matrix $\bfA \in \R^{\dimxdim}$ (assumed to be full-rank throughout this work). Given any factorization $\bfA = \bfB\bfC$, a DP estimate of $\bfA\obsM$ can be computed as
\begin{equation}\label{eq:mat_mech_def}
\widehat{\bfA\obsM} = \bfB\left(\bfC\obsM + \noiseM\right)
\end{equation}
where $\noiseM$ represents a sample from a noise distribution $\mathcal{D}$. Choosing $\mathcal{D}$ in a way that appropriately depends on the sensitivity of the map $\obsM \mapsto \bfC\obsM$, one can prove privacy of the noised vector $\bfC\obsM + \noiseM$, and hence of the mapping $\obsM \mapsto \widehat{\bfA\obsM}$. For example, in the case of cumulative sums, the matrix $\bfA$ is the lower-triangular matrix $\bfS$ with 1's on and below the diagonal; the binary tree mechanisms correspond to a matrix $\Ctree$ with one row per tree node (see \cref{app:special_matrices} and \cref{app:tree_agg_matfac} respectively).  A typical choice for the noise $\noiseM$ is to select it from a spherical Gaussian distribution.

A focus of existing work (e.g. \cite{Li2015TheMM,opt_convex_fact,hdmm,edmonds_nikolov_ullman}) is to choose the factorization $\bfB\bfC$ to optimize some measure of overall accuracy (such as total mean squared error) subject to a privacy constraint. However (with the exception of the independent, parallel work of \citet{fichtenberger_constant22}\footnote{\citet{fichtenberger_constant22} strive to a get an analytical optimal leading multiplicative constant for the additive error achievable for the problem of continual observation under DP. We on the other hand focus on computationally estimating the optimal matrix factorization mechanism for the problem under DP.  We leave the empirical comparison to the explicit construction in  Fichtenberger et al. for future work.}), streaming constraints and adaptive privacy were not explicitly considered. 
In the streaming setting, one naturally requires the $i^{th}$ element (row) of $\bfA\obsM$ to be computable using only the first $i$ elements (rows) of $\obsM$. This corresponds to requiring that the linear operator of interest $\bfA$ has a lower-triangular structure when represented as a matrix, a requirement that is met by all the matrices under consideration in this work. 
Distributing the $\bfB$ in \cref{eq:mat_mech_def}, we see $\widehat{\bfA\obsM} = \bfA\obsM + \bfB\bfZ$. As long as $\bfA\obsM$ is lower triangular, then, \emph{any} matrix mechanism can be implemented by an online algorithm, via the distribution induced by $\bfB\noiseM$. However, this observation does not address a key problem: under what conditions is the resulting mechanism \textit{adaptively} private? 

The applications to optimization raise their own set of critical questions: How can we efficiently compute such factorizations? Which linear operators are the appropriate ones to factorize in the case of SGD? Finally, can these factorizations actually produce improved privacy/accuracy tradeoffs for real-world machine learning tasks?

\begin{figure}[t]
    \centering
    \makebox[\textwidth][c]{\includegraphics[width=1\textwidth]{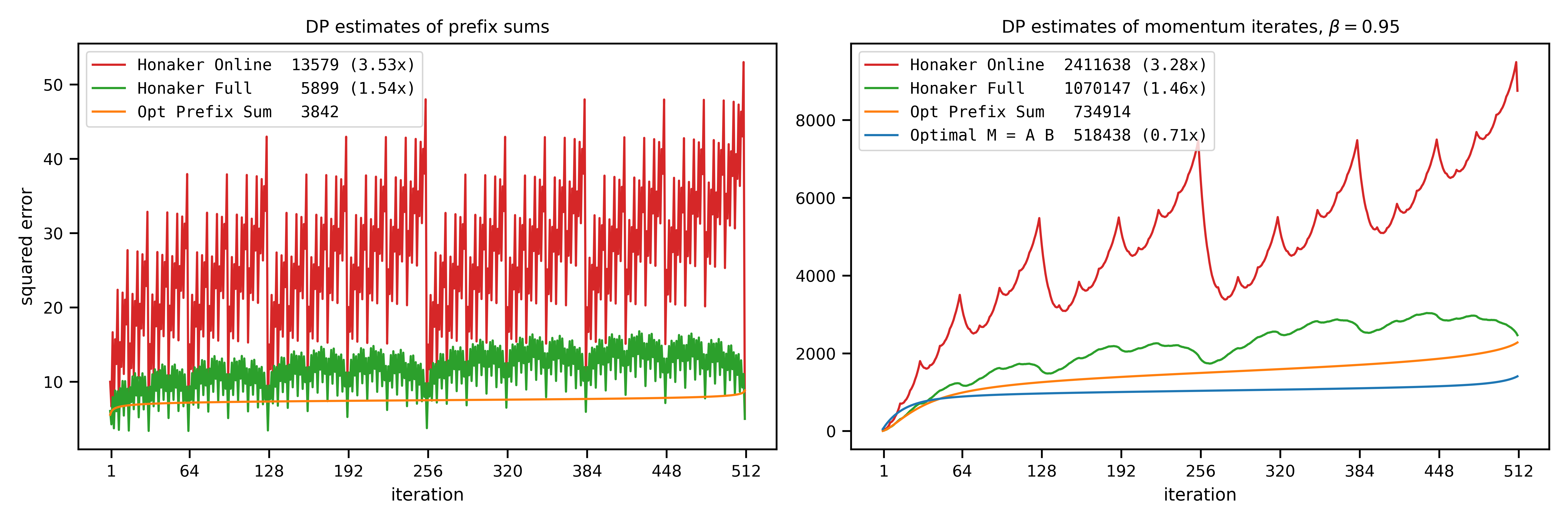}}%
    \vspace{-0.3in}
    \caption{Left: Per-iteration squared error of three mechanisms for DP online prefix sums for $\dimension=512$.  \HonakerOnline and \HonakerFull correspond to DP binary tree aggregation with the streaming and full Honaker estimators (respectively), while \OptPrefixSum corresponds to the optimal matrix factorization. The tree-based mechanisms suffer from variability in the error due to the binary tree structure. Right: Squared error in the DP estimates of the iterates of momentum SGD for four mechanisms. Momentum is treated as post-processing of cumulative sums for the first 3 mechanisms,  while \OptMomentum uses the optimal factorization of the momentum matrix (see \cref{sec:more_mechanisms}).}
    \label{fig:momentum_error}
\end{figure}

\mypar{Contributions} 
We provide a deeper understanding of the matrix mechanism in the adaptive streaming setting. We show that if $\noiseM$ in \cref{eq:mat_mech_def} is drawn from a Gaussian distribution of appropriately computed variance, then the resulting mechanism is differentially private in the adaptive streaming setting, \emph{independent of the structure of $\bfB$ and $\bfC$}. We make an explicit connection here with the easier-to-see privacy under adaptive streams of lower-triangular factorizations. Furthermore, we show that this property is specific to the Gaussian mechanism, and does not extend to arbitrary noise distributions. %

For natural notions of error and adjacency specified in \cref{sec:computing_optima}, we present a fast and parameter-free fixed-point algorithm for computing optimal factorizations and prove a local convergence guarantee for this algorithm, leveraging representations of optimal factorizations which are to our knowledge novel in the literature on the matrix mechanism. The optimal computed factorizations show a significant improvement over existing state-of-the-art private streaming prefix sum methods~\cite{honaker_trick}, additionally removing the artifacts of the binary tree data structure (see \cref{fig:momentum_error}). Furthermore, our fixed-point algorithm can be two orders of magnitude more computationally efficient than existing optimization methods for the matrix mechanism~\cite{opt_convex_fact} (though we did not explicitly compare to~\citet{hdmm}), and provides a direct bound on the duality gap which allows precise stopping criteria.

Going beyond prefix sums (which correspond to constant learning rate SGD as noted above), we construct matrix mechanisms that directly encode more sophisticated optimization algorithms as linear operators on gradients: in particular, arbitrary combinations of (data independent) learning rate schedules and momentum.

We compute optimal factorizations of these general matrices via fixed-point iterations, and use them to train user-level differentially private language models on a canonical federated learning benchmark, showing that these factorizations  significantly improve the privacy/utility curve (in fact, closing 2/3rds of the gap to non-private training left by the previous state-of-the-art for single pass algorithms). For prefix sums, we show computationally-efficient structured matrices provide high-fidelity approximations to the optimal matrices, allowing implementations to scale essentially independent of the number of iterations $\dimension$.

\mypar{Notation and conventions}
Matrices will be denoted by bolded capital letters (e.g. $\bfA$, $\bfB$), with some symbols reserved for special matrices, notably $\bfS$ (prefix sums) and  $\bfM$ (momentum, defined in \cref{sec:more_mechanisms} and illustrated in \cref{app:special_matrices}). Vectors will be denoted by bolded lowercase letters (e.g. $\bfx$, $\bfy$). For a real symmetric matrix $\bfA$, the smallest and the largest eigenvalues are denoted by $\lambda_{\min}(\bfA)$ and $\lambda_{\max}(\bfA)$. For a matrix $\bfA$, $\bfA^*$ denotes the conjugate transpose and $\bfA^\dagger$ denotes the Moore-Penrose pseudoinverse; a star as in $\bfX^\star$ indicates a matrix that is optimal in a way made clear by context. $\diagpart$ represents the operation of taking the diagonal elements of a matrix; $\diag$ represents embedding a vector argument on the diagonal of a matrix. Additional notation is summarized in \cref{app:special_matrices}.

\section{Privacy for adaptive streams}
\label{sec:streaming_priv}

In this section, we provide structural results that clarify the classes of mechanisms/algorithms for cumulative sums and other linear computations in the \emph{continual release model}~\citep{Dwork-continual,CSS11-continual} that remain private even when the inputs stream is defined \textit{adaptively}. In the continual release model, a mechanism receives a stream of inputs $\obsM =[\obs_1,\ldots, \obs_\dimension]$ and produces a stream of outputs $\bfa_1,...,\bfa_\dimension$, where output $\bfa_i$ is intended to approximate some function of the prefix $\obs_1,...,\obs_i$ and must be generated before $\obs_{i+1}$ is received.  
We specify which parts of the input can depend on a single person's data via a \textit{neighbor} relation $\cN$ on data streams. Two streams are neighbors if they differ in one person's data. For example, if  one person's data directly affects exactly one input in the stream (``\textit{event-level privacy}''), then we say two data streams are \textit{neighboring} if they differ in exactly one element (that is, they are at Hamming distance 1). 

The original works on continual release analyzed privacy in a \textit{nonadaptive} model: a mechanism $\mech$ is $(\eps,\delta)$-differentially private~\cite{DMNS,ODO} for nonadaptive continual release if, for all pairs of adjacent data streams $\obsM, \obsMp$, the corresponding distributions on output streams $\mech(\obsM)$ and $\mech(\obsMp)$ are $(\eps,\delta)$-indistinguishable, denoted $\mech(\obsM) \approx_{\eps,\delta}\mech(\obsMp)$. That is, for all events $E$, we have $\Pr[\mech(\obsM) \in E] \leq e^\eps \Pr(\mech(\obsMp) \in E) + \delta$ and  $\Pr(\mech(\obsMp) \in E) \leq e^\eps \Pr(\mech(\obsM) \in E) + \delta$. (A variant of this definition tailored to differentially private gradient descent, with a precise instantiation of the neighborhood notion is presented in Definition~\ref{def:diiffP} in the appendix.)

\newcommand{\cmpl}[1]{\tilde{#1}}  %
In many use cases---including those arising in iterative gradient-based optimization algorithms---the nonadaptive model is inadequate, since the inputs $\obs_i$ may be generated in real time as a function of previous outputs $\bfa_1,...,\bfa_{i-1}$. To summarize the more general, adaptive definition \citep{SmithThakurta13,JainRSS21}, consider an adversary that \textit{adaptively} defines two input sequences $\obsM= (\obs_1,\dots, \obs_\dimension)$ and $\obsMp=(\obsp_1,\dots, \obsp_\dimension)$. The adversary must satisfy the promise that these sequences correspond to neighboring data sets. 
The privacy game proceeds in rounds. At round $t$, the adversary generates $\obs_t$ and $\obsp_t$. 
The game accepts these if the input streams defined so far are valid, meaning that there exist completions $(\cmpl{\obs}_{t+1},...,\cmpl{\obs}_\dimension)$ and $(\cmpl{\obsp}_{t+1},...,\cmpl{\obsp}_\dimension)$ so that 
$
((\obs_1,...,\obs_t,\cmpl{\obs}_{t+1},...,\cmpl{\obs}_\dimension), 
(\obsp_1,...,\obsp_t,\cmpl{\obsp}_{t+1},...,\cmpl{\obsp}_\dimension)) \in  \cN\, .
$
For example, in the case of event-level privacy, the game simply checks that the two streams differ in at most one position so far. 

The game is parameterized by a bit $\side \in \bit{}$ which is unknown to the adversary but constant throughout the game. The game hands either $\obs_t$ or $\obsp_t$ to the mechanism $\mech$, depending on $\side$. The mechanism's output $\bfa_t$ is then sent to the adversary. The privacy requirement is that the adversary's views with $\side =0$ and $\side=1$ be $(\epsilon,\delta)$ indistinguishable.  
One can substitute other relevant notions of indistinguishability like those from CDP~\citep{bun2016concentrated,DworkR16}, Renyi DP~\citep{mironov2017renyi}, or Gaussian DP \cite{GDP-DL}. The mechanisms we consider generally satisfy Gaussian DP.

The nonadaptive version of the definition is weaker but easier to work with. It is therefore natural to look for classes of mechanisms for which the two definitions are equivalent (and thus for which a nonadaptive privacy proof implies the more general guarantee). 
We first observe that such a transfer statement holds for ``pure'' $\epsilon$-DP (in which $\delta=0$). We defer all the proofs to Appendix~\ref{app:streaming_priv}.

\begin{prop}
Every mechanism that is $(\epsilon,0)$ nonadaptively DP in the continual release model satisfies the \textit{adaptive} version of the definition, with the same parameters.
\label{prop:adaptiveDPpure}
\end{prop}

Unfortunately, not all privacy proofs for the nonadaptive model transfer to the adaptive setting. Indeed, we show that there are additive-noise mechanisms (which simply add noise from a pre-defined distribution to some function of the data) that are nonadaptively $(\eps,\delta)$-DP, but \textit{not} private in the adaptive setting (Appendix~\ref{sec:nonadaptivePrivNonGaussian}). %

\paragraph{Adaptive privacy for Gaussian noise addition mechanisms}
In Theorem~\ref{thm:GaussianAdaptive} we show that every matrix mechanism with Gaussian noise addition that is $(\epsilon,\delta)$-DP in the nonadaptive model is also private in the adaptive setting:

\begin{thm}\label{thm:GaussianAdaptive}
Let $\bfA\in\mathbb{R}^{\dimension\times \dimension}$ be a lower-triangular full-rank query matrix, and let $\bfA=\bfB\bfC$ be any factorization with the following property: for any two \textit{neighboring} streams of vectors $\bfG\,, \bfH\in \R^\datadim$, we have $\|\bfC(\bfG-\bfH)\|_F\leq \kappa$. Let 
$\bfZ\sim\calN(0,\kappa^2\sigma^2)^{\dimension\times d}$ with $\sigma$ large enough so that $\mech(\bfG) = \bfA\obsM + \bfB\bfZ = \bfB(\bfC \obsM +\bfZ)$ satisfies 
$(\epsilon,\delta)$-DP (or $\rho$-zCDP or $\mu$-Gaussian DP) in the \textit{nonadaptive} continual release model. Then, $\mech$ satisfies the same DP guarantee (with the same parameters) even when the 
rows of the input are chosen adaptively.
\end{thm}

To prove this we crucially use the rotational invariance of spherical Gaussian distribution, yielding distributional equivalence of an orbit of mechanisms: those factorizations expressible as $\bfB\bfU\bfU^*\bfC$ for $\bfU$ unitary.  This observation can similarly be leveraged to show a subtly distinct fact:
\begin{prop}\label{prop:lt_factorizations} For any factorization $\bfA=\bfB\bfC$ where $\bfA$ is lower-triangular, there exists a factorization $\bfA=\widehat{\bfB}\widehat{\bfC}$ which induces a distributionally equivalent matrix mechanism under Gaussian noise with $\widehat{\bfB}$ and $\widehat{\bfC}$ lower triangular. This factorization can be explicitly computed from $\bfA=\bfB\bfC$ via an appropriate LQ decomposition. Normalizing to all-nonnegative entries on the diagonal, this factorization is unique.
\end{prop}

\section{Computing optimal factorizations}\label{sec:computing_optima}

Once one writes down the matrix mechanism as in \cref{eq:mat_mech_def}, it is natural to seek factorizations that minimize the error $\widehat{\bfA\obsM} - \bfA \obsM$ in some metric of choice. In this section we consider the expected squared reconstruction error of the estimate $\widehat{\bfA\obsM}$, which has previously been noted as an appropriate formulation of error for the setting of training private ML models \citep[Theorem 5]{kairouz2021practical}. Further, for simplicity we restrict our attention to the single-pass setting. That is, for the remainder of the paper we will assume:

\begin{definition}\label{assumption:single_pass}
Two data matrices $\obsM$ and $\obsMp$ in $\R^\datadim$ will be considered to be neighboring if they differ by a single row, with the $\ell_2$-norm of the difference in this row at most $\clipnorm$.
\end{definition}

Under this notion of sensitivity, one can make  the estimation of any query $\bfA\obsM$ $(\epsilon,\delta)$-DP via  Theorem~\ref{thm:priv1}, which via \cref{thm:GaussianAdaptive} immediately extends to adaptive streams. It is worth mentioning that while we state the analytic vairance for the Normal distribution to satisfy $(\epsilon,\delta)$-DP, in practice we arrive at the required (and tighter) variance via empirical privacy accounting methods like zero concentrated DP (zCDP) accounting~\cite{bun2016concentrated}, or privacy loss distribution (PLD) accounting~\cite{koskela2021tight}.

\begin{thm}[Adapted from~\cite{Li2015TheMM}]\label{thm:matmech_privacy_quant}
Consider a query matrix $\bfA\in\mathbb{R}^{\dimensions\times\dimensions}$ along with a fixed factorization $\bfA=\bfB_{\dimension\times\dimension}\bfC_{\dimension\times\dimension}$ with $\gamma=\max_{i\in[\dimensions]}\ltwo{\bfC\idx{:}{i}}$, the maximum column norm of $\bfC$. Let $\obsM\in\mathbb{R}^{\datadim}$ be a fixed (non-adaptive) data matrix with each row of $\obsM$ having $\ell_2$-norm at most $\zeta$. The algorithm that outputs $\bfB\left(\bfC\obsM+\bfZ\right)$ with $\bfZ\sim\mathcal{N}\left(0,\frac{\gamma^2\zeta^2\left(2\log(1/\delta)+\epsilon\right)}{\epsilon^2}\right)^{\dimension\times \mdim}$ satisfies $(\epsilon,\delta)$-DP.
\label{thm:priv1}
\end{thm}

In this setting, with $\bfZ \sim \mathcal{N}\left(0, \gamma^2 \right)^{\dimension\times \mdim}$ following Theorem~\ref{thm:matmech_privacy_quant} (which ensures a fixed level of privacy when $\clipnorm=1$ for an arbitrary factorization factorization $\bfA = \bfB\bfC$) the expected reconstruction error can be computed directly as $\calL(\bfB, \bfC)$ \citep[Proposition 9]{Li2015TheMM}, \citep[Equation 3]{opt_convex_fact}, 
\begin{equation}\label{eq:l_expression}
    \gamma^2(\bfC) = \max_{i \in [1, \dots, \dimension]} \left\|\bfC_{[:, i]}\right\|_2^2
    \qquad \text{and} \qquad
    \calL(\bfB, \bfC) = \gamma^2(\bfC) \left\|\bfB\right\|_F^2.
\end{equation}

As has been noted \citep{Li2015TheMM,opt_convex_fact}, \cref{eq:l_expression} can be manipulated to yield a convex program, for which hand-tuned algorithms exist \citep[Section 4]{opt_convex_fact}. \cref{thm:GaussianAdaptive} shows, for the first time, that arbitrary factorizations found by minimizing this optimization problem can be applied in the adaptive streaming setting. 

We present an alternative characterization of these optima, which reformulates the optimization problem as a fixed-point problem. We show that simply iterating an explicit mapping converges to this fixed point from an appropriate initialization, and observe numerically that the associated algorithm achieves fast, global convergence.%

Since the Moore-Penrose pseudoinverse yields the minimal $\ell_2$-norm solution to a set of underdetermined linear equations \citep[Theorem 2.1.1]{nla.cat-vn1139431}, we note that for a fixed $\bfC$ term (of any dimensionality), the optimal $\bfB$ may be expressed as $\bfB_{\bfC}^\star = \bfA\bfC^\dagger$. Since $\bfA = \bfB\bfC$ implies $\bfA = \big(\alpha\bfB\big)\big(\frac{1}{\alpha}\bfC\big)$, for any linear space of matrices $\bfV$, we may express the optimization problem of interest as

\begin{equation}\label{eq:constrained_problem}
\min_{\bfC \in \bfV} \calL\left(\bfA\bfC^\dagger, \bfC\right) 
    = \min_{\bfC \in \bfV} \gamma^2(\bfC) \left\|\bfA\bfC^\dagger\right\|_F^2 
    = \min_{\bfC \in \bfV, \gamma^2(\bfC)=1} \left\|\bfA\bfC^\dagger\right\|_F^2.
\end{equation}

The properties of the problem \cref{eq:constrained_problem} have been studied previously. In particular, \citep[Section 3]{opt_convex_fact} studied a symmetric version, transforming the problem as:

\begin{equation}\label{eq:symmetrized_problem}
    \xopt = \argmin_{\bfX \text{ is PD}, \bfX\idx{i}{i} \leq 1, 1 \leq i \leq n} \tr(\bfA^* \bfA \bfX^{-1})
\end{equation}

which essentially reparameterizes \cref{eq:constrained_problem} (with $\bfV = \R^\dimension$) in terms of $\bfC^*\bfC$. To recover a matrix-mechanism factorization of $\bfA$, then, one may utilize any $\bfC$ such that $\xopt = \bfC^*\bfC$, e.g. $\bfC = \sqrt{\xopt}$. \cref{prop:lt_factorizations} can be used to construct a lower-triangular factorization if desired. 

\citet{opt_convex_fact} show:
1) Any solution $\xopt$ of \cref{eq:symmetrized_problem} must have diagonal entries exactly 1. 
2) Any solution $\xopt$ may be taken to be strictly within the positive-definite cone, with minimal eigenvalue bounded from below in terms of the eigenvalues of $\bfA$.
3) For any full-rank $\bfA$, $\bfX \mapsto \tr(\bfA^* \bfA \bfX^{-1})$ is strictly convex over symmetric, positive-definite matrices. Therefore the solution to \cref{eq:symmetrized_problem} is unique.

By analyzing \cref{eq:symmetrized_problem} directly, we derive a characterization of solutions in terms of an explicit fixed-point problem, with a corresponding bound on the optimality gap.

\begin{thm}\label{thm:expressions_for_x}
The minimizer $\xopt$ of \cref{eq:symmetrized_problem} is in one-to-one correspondence with the unique fixed point of the function $\phi: \R^n_+ \to \R^n_+$ defined by
\begin{equation}\label{eq:phi_def}
\phi(\bfv) = \diagpart\left(\sqrt{\diag(\bfv)^{1/2} \,\bfA^*\bfA\,\diag(\bfv)^{1/2}}\right).
\end{equation}
Letting
\begin{equation}\label{eq:x_for_dualv}
  \Xofv(\dualv) = \diag(\dualv)^{-1/2} \left(\diag(\dualv)^{1/2}\,\bfA^*\bfA\,\diag(\dualv)^{1/2}\right)^{1/2} \diag(\dualv)^{-1/2},
\end{equation}
for the fixed point $\vopt$ of \cref{eq:phi_def}, that is $\phi(\vopt) = \vopt$, we have $\xopt = \Xofv(\vopt)$, and this pair $\left(\xopt, \vopt\right)$ satisfies
\begin{equation}\label{eq:lagrangian_zero_def}
  \bfA^* \bfA = \xopt\diag\left(\vopt\right)\xopt.
\end{equation}
Further, for any $\dualv \in \R^\dimension_+$, the objective value of the primal problem \cref{eq:symmetrized_problem} is lower-bounded by
\begin{equation}\label{eq:lower_bound_from_dual}
\tr\big(\diag(\dualv) (2\Xofv(\dualv) - \bfI)\big),
\end{equation}
and this bound is tight for $\dualv = \vopt$.
\end{thm}

The sum of the elements of $\phi$ represents a quantity of independent interest in quantum information, the so-called Jozsa fidelity \citep{quantum_fidelity_def,Liang_2019}, while $\Xofv$ represents the matrix geometric mean of $\bfA^*\bfA$ and $\diag(\dualv)^{-1}$ \citep{matrix_geom_mean,positive_linop_means}. These connections give some hope that the fixed point of $\phi$ can be understood in a direct manner. 
 We can show local, though not yet global, convergence of the iterates of $\phi$ to this fixed point.

\begin{thm}\label{thm:local_contraction}
$\phi$ defined in \cref{eq:phi_def} is a local contraction around its fixed point in a suitable metric, and hence there is a neighborhood of this fixed point in which iterates of $\phi$ converge to this fixed point. The precise norm of this contraction, and the size of the neighborhood in which convergence is guaranteed, can both be estimated in terms of minimum and maximum eigenvalues of $\bfA$.
\end{thm}

This result can be shown by linearizing the mapping $\phi$ around its fixed point and performing an involved estimate of its Jacobian at the fixed point. As such, it is implied by \cref{sd4}, which states that the linearization of $\phi$ around its fixed point is a contraction in a suitable metric, and therefore the Banach fixed-point theorem applies.

\textbf{Remark.} Notably missing from quantification of the contraction is the dimension $\dimension$. Indeed, the argument is dimension-independent in a strong sense: it applies to the suitably generalized definition of $\phi$ where $\bfA$ is any bounded linear operator on a Hilbert space with bounded inverse.

\paragraph{Empirical performance of the fixed-point method}
Experimentally, iterating the mapping $\phi$ is sufficient to converge to the global optimum extremely quickly from any initial point (modulo potential numerical issues in computing the matrix square root, discussed in \cref{app:numerics_of_phi}).
\citet[Algorithm 1]{opt_convex_fact} design an algorithm with globally linear and locally quadratic convergence rate, with similar asymptotics to iterating $\phi$ (each dominated by an $\dimension^3$ term), though at the cost of introducing a parameter $T$. Iterating $\phi$, on the other hand, is parameter-free.
We implemented \citep[Algorithm 1]{opt_convex_fact} as well as a direct gradient-descent-based method to numerically compare convergence speed. As a canonical benchmark, we computed optimal factorizations of the $512\times512$ and $2048\times2048$ prefix-sum matrices $\bfS$ (\cref{sec:optfactfig} provides a visualization of this optimal factorization). Our fixed-point algorithm was significantly faster than either of the alternatives to compute optima, computing a lower-loss matrix for the larger problem in less than 3 minutes than either alternative found in over 80 minutes. See \cref{app:numerics_of_phi} for details.  Further, via \cref{eq:lower_bound_from_dual}, our approach provides an optimality certificate that allows a precise specification of the stopping criteria in terms of any target optimality gap. The speed and simplicity of our fixed-point algorithm was a significant enabler of the mechanism exploration presented in the next section. 

\section{The matrix mechanism for SGD}\label{sec:more_mechanisms}
To define our gradient descent algorithm, let $\obsM \in \R^\datadim$ be the matrix of gradients, with row vector $\obs_i \in \R^{1 \times \mdim}$ the gradient observed on iteration $i$ after clipping to norm at most $\clipnorm$; we abuse notation slightly by writing $\obsMti \in \R^\datadim$, formed by taking the first $i$ rows of $\obsM$, with zeros for the as-of-yet unobserved gradient rows for iterations $i+1, \dots, \dimension$ (the lower triangular structure of the matrices we consider will imply $\obsMti$ vs $\obsM$ does not in fact change the value computed). With this notation, we define \cref{alg:dpmfsgd}, a general template for private SGD algorithms.
\newcommand{\ghat}{\hat{\bfg}}
\begin{figure}
\begin{minipage}[t]{0.45\linewidth}
\centering
    \begin{algorithm}[H]
    \caption{DP Matrix Factorization SGD}
    \label{alg:dpmfsgd}
    \begin{algorithmic}[1]
    \State Inputs: 
    \State $\quad$ factorization $\bfM = \bfB \bfC$
    \State $\quad$ overall learning rate $\eta$
    \State $\quad$ noise level $\sigma$, clipping norm $\clipnorm$
        \State $\quad$ examples $\example_i$, $i \in \{1, \dots, \dimension\}$
    
    \State $\bftheta\idx{0}{:} \assign 0 \in \R^\mdim$
    \State Sample  $\bfZ \in \R^\datadim$,  $\bfZ\idx{i}{j}\!\sim\!\calN(0,\sigma^2)$ iid
    \For{$i$ in $1, \dots, \dimension$} 
      \State  $\ghat \assign \nabla_{\bftheta\idx{i-1}{:}} \ell(\bftheta; \example_i)$
      \State $\obsM\idx{i}{:} \assign \ghat \cdot \min\left\{\frac{\clipnorm}{\|\ghat\|_2}, 1\right\}$      
      \State $\bftheta\idx{i}{:} \assign -\eta \big(\bfM\idx{i}{:} \obsM\idx{1:i}{:} + \bfB\idx{i}{:} \noiseM\big)$
    \EndFor
    \end{algorithmic}
    \end{algorithm}
\end{minipage}
\hfill
\begin{minipage}[t]{0.45\linewidth}
\centering
    \begin{algorithm}[H]
    \caption{Heavy-ball momentum}
    \label{alg:momentum}
    \begin{algorithmic}[1]
    \State $\bftheta_0 \assign 0$, $\bfm_0 \assign 0$
    \For{$i$ in $1, \dots, \dimension$} 
      \State $\bfm_i \assign \beta \cdot \bfm_{i-1} + \obs_i$
      \State $\bftheta_i \assign \bftheta_{i-1} - \eta_i \bfm_i$
    \EndFor
    \end{algorithmic}
    \end{algorithm}
\end{minipage}
\end{figure}
The power of this general formulation comes largely from the following privacy guarantee:
\begin{thm}\label{thm:dpmfsgd}
 Under the ``replace with zero'' notion of differential privacy (in Defininition~\ref{def:diiffP} in the appendix) over examples (records) $\example_i$, taking $\ghat = 0$ when $\example_i = \bot$, \cref{alg:dpmfsgd} (that releases the iterates $\bftheta\idx{i}{:}$) satisfies equivalent $(\eps, \delta)$-DP to the Gaussian mechanism with noise variance $\sigma^2$ applied to records with $\ell_2$ sensitivity at most $\clipnorm \gamma$, where $\gamma = \max_{i} \left\|\bfC_{[:, i]}\right\|_2$ is the maximum column norm of $\bfC$, and $\zeta$ is the clipping norm.
\end{thm}

\cref{thm:dpmfsgd} shows that the contribution of $\bfC$ to the loss~\cref{eq:l_def} is reflected in the privacy guarantee of~\cref{alg:dpmfsgd}, by determining the sensitivity of the matrix mechanism~\cref{eq:mat_mech_def}. The contribution of $\bfB$ determines the expected squared reconstruction error of the matrix mechanism, by definition. This quantification can in turn be converted by existing analytical methods to a regret bound for convex losses:

\begin{prop}[Adaptation of Theorem C.1,~\citet{kairouz2021practical}]\label{prop:sgd_utility}

In the setup of~\cref{alg:dpmfsgd}, let $\bfM$ be the prefix-sum matrix $\bfS$, and assume $\ell$ is convex with $\ell_2$-Lipschitz constant $L$. Let $\theta_t = \bftheta_{[t, :]}$. For any $\thetaopt \in \R^\mdim$, 

$$
\frac{1}{\dimension}\sum_{t=1}^n\mathbb{E}\left[\ell(\theta_t;\chi_t)-\ell(\thetaopt;\chi_t)\right] \leq 
\eta L^2 + \frac{1}{2 \eta \dimension}\left(\|\thetaopt\|_2^2 - \|\theta_1\|_2^2\right) + \frac{L\sigma\eta}{\sqrt{\dimension}}\|\bfB\|_F
$$

\end{prop}

We now consider different instantiations of \cref{alg:dpmfsgd}. Let $\bfS$ be the prefix-sum matrix as defined in \cref{app:special_matrices}, and let $\Ctree$ be the matrix representation of the binary tree (\cref{app:tree_agg_matfac}), so for appropriate choices of the reconstruction matrices $\Bhs$ and $\Bhf$,  $\bfS = \Bhs \Ctree$ gives the \HonakerOnline mehcanism, and $\bfS = \Bhf \Ctree$ gives the \HonakerFull mechanism. In particular, using the \HonakerOnline factorization in \cref{alg:dpmfsgd} recovers the non-momentum DP-FTRL algorithm of \citet{kairouz2021practical}.

However, \citet{kairouz2021practical} observed that for non-convex objectives, DP-FTRL with momentum provided superior privacy/accuracy tradeoffs. Given prefix sums, momentum can be implemented as post processing by estimating individual gradients/updates as the difference of successive cumulative sums (multiplication by $\bfS^{-1}$), and then passing these into a standard momentum SGD optimizer. We show that performance can be improved by directly incorporating momentum and a-priori learning rate schedules directly into the DP mechanism.

A basic but important observation is that momentum SGD can be expressed as a linear map of gradients $\obsM \to \bfM\obsM$. We consider the classic momentum algorithm of \citet{polyak64}, with per-iteration learning rates\footnote{The schedule may be arbitrary, but must be chosen a priori in a data-independent way.} $\eta_1, \dots, \eta_n$ and momentum $\beta \in [0, 1)$, as in \cref{alg:momentum}.
Alternatively, we can express momentum SGD as a linear operator on the gradients:  
\begin{prop}\label{prop:momentummatrix}
For any $\beta \in [0, 1)$, $\dimension \ge 1$, per iteration learning rates $\eta_1, \dots, \eta_\dimension$, define the lower-triangular matrix $\bfM \in \R^{\dimension \times \dimension}$ as the product of lower-triangular matrices $\ML$ and $\MR$:
\begin{equation}
\ML\idx{i}{j} = 
    \begin{cases}
       \eta_j   & \text{$i \ge j$} \\
       0        & \text{otherwise}
    \end{cases},
\quad
\MR\idx{i}{j} = 
    \begin{cases}
       \beta^{i - j} & \text{$i \ge j$} \\
       0             & \text{otherwise}
    \end{cases},
\qquad
\text{and}
\qquad
\bfM = \ML \MR.
\end{equation}
Then, for any matrix of of per-iteration gradients $\obsM \in \R^\datadim$ with rows $[\obs_1, \dots, \obs_\dimension] $, the sequence of iterates $\bftheta \in \R^\datadim$ with rows $[\bftheta_1, \dots \bftheta_\dimension]$ produced by \cref{alg:momentum} can equivalently be written $\bftheta = -\bfM \obsM$.
\end{prop}

We apply the fixed-point algorithm of \cref{sec:computing_optima} to $\bfM$ to obtain optimal matrix mechanisms $\bfM = \bfB \bfC$, which indeed leads to improved performance.\footnote{The definition $\bfM = \ML \MR$ is for the convenience, and is unrelated to the optimal factorization.} 
We can also convert mechanisms that produce DP prefix sums to produce momentum iterates via post processing: given any matrix mechanism for the prefix sum problem given as a factorization $\bfS = \bfB \bfC$, we can convert this to a mechanism for momentum as $\bfM = \hat{\bfB} \bfC$ where $\hat{\bfB} = \bfM \bfS^{-1} \bfB$. Straightforward calculations show this representation is equivalent to the ``Momentum Variant'' of DP-FTRL \citep{kairouz2021practical}. This allows us to consistently evaluate the mechanisms in terms of the total variance (squared error) induced in the outputs by unit-variance noise $\noiseM$ in the mechanism via \cref{eq:l_expression}.
\cref{tab:algs} summarizes the four instantiations of \cref{alg:dpmfsgd} for any choise of $\bfM$; note in particular that for $\beta=0$ and a fixed learning rate schedule $\eta_i = 1$, $\bfM = \bfS$.
\cref{fig:momentum_error} compares the per-step squared error of  the DP momentum iterates for these methods for $\beta=0$ (prefix sums, as a constant $\eta=1$ is used) and $\beta=0.95$.

\begin{table}
\renewcommand{\arraystretch}{1.3}
\begin{tabular}{llll}
Mechanism        & $\bfB$                    & $\bfC$ &  s.t. $\bfM = \bfB \bfC$\\
\hline                 
\HonakerOnline   & $\bfM \bfS^{-1} \Bhs$     & $\Ctree$ & Equivalent to DP-FTRL of \citet{kairouz2021practical} \\
\HonakerFull     & $\bfM \bfS^{-1} \Bhf$     & $\Ctree$ & \\
\OptPrefixSum    & $\bfM \bfS^{-1} \bfB_S^*$ & $\bfC_S^\star$ & for optimal $\bfS = \bfB_S^\star \bfC_S^\star$ \\
\OptMomentum     & $\bfB_M^\star$            & $\bfC^\star_M$ & for optimal $\bfM = \bfB_M^\star \bfC_M^\star$ \\
\hline \\
\end{tabular}
\caption{Instantiations of \cref{alg:dpmfsgd} for various factorizations of the SGD matrix $\bfM = \bfB \bfC$.} \label{tab:algs}
\end{table}

\mypar{Computational efficiency}
Some care is necessary for the efficient implementation of \cref{alg:dpmfsgd}, particularly the computation of $\bfM\idx{i}{:} \obsM\idx{1:i}{j} + \bfB\idx{i}{:} \bfZ$ on line 11. First, we observe that via \cref{prop:momentummatrix} we can efficiently compute the $\bfM \obsM$ term via \cref{alg:momentum}, rather than as a matrix operation. This leaves the computation of the noise $\bfB\idx{i}{:} \bfZ.$ For applications to ML, $\mdim$ could be $10^6 - 10^9$, and with even $\dimension = 10^4$ rounds (iterations) this might make the total calculation quite expensive if not prohibitive. With an efficient TensorFlow implementation, in our experiments with $\mdim \approx 4 \times 10^6$ and $\dimension=2048$, we found we could compute the noise directly. However, for larger applications we show (in \cref{sec:efficient}) that one can compute structured matrices that well-approximate the optimal $\bfB^\star$ for the prefix sum matrix while allowing for $\mathcal{O}(\mdim)$ calculation of the per-round noise vectors (on par with that of the binary tree mechanism, which can also provide computational efficiency with a careful implementation, see \cref{tab:loss_values}).  The key is to observe the diagonal dominance of the optimal $\bfB^\star$ (see \cref{sec:optfactfig}), leading to an approximation $\hat{\bfB}$ that is the sum of a lower-triangular $d$-banded matrix with the remaining entries in the lower triangle extracted from a low-rank approximation computed via alternating-least-squares. 

\begin{figure}[!t]
\begin{minipage}[t]{0.48\linewidth}
    \centering
    \includegraphics[width=1\linewidth]{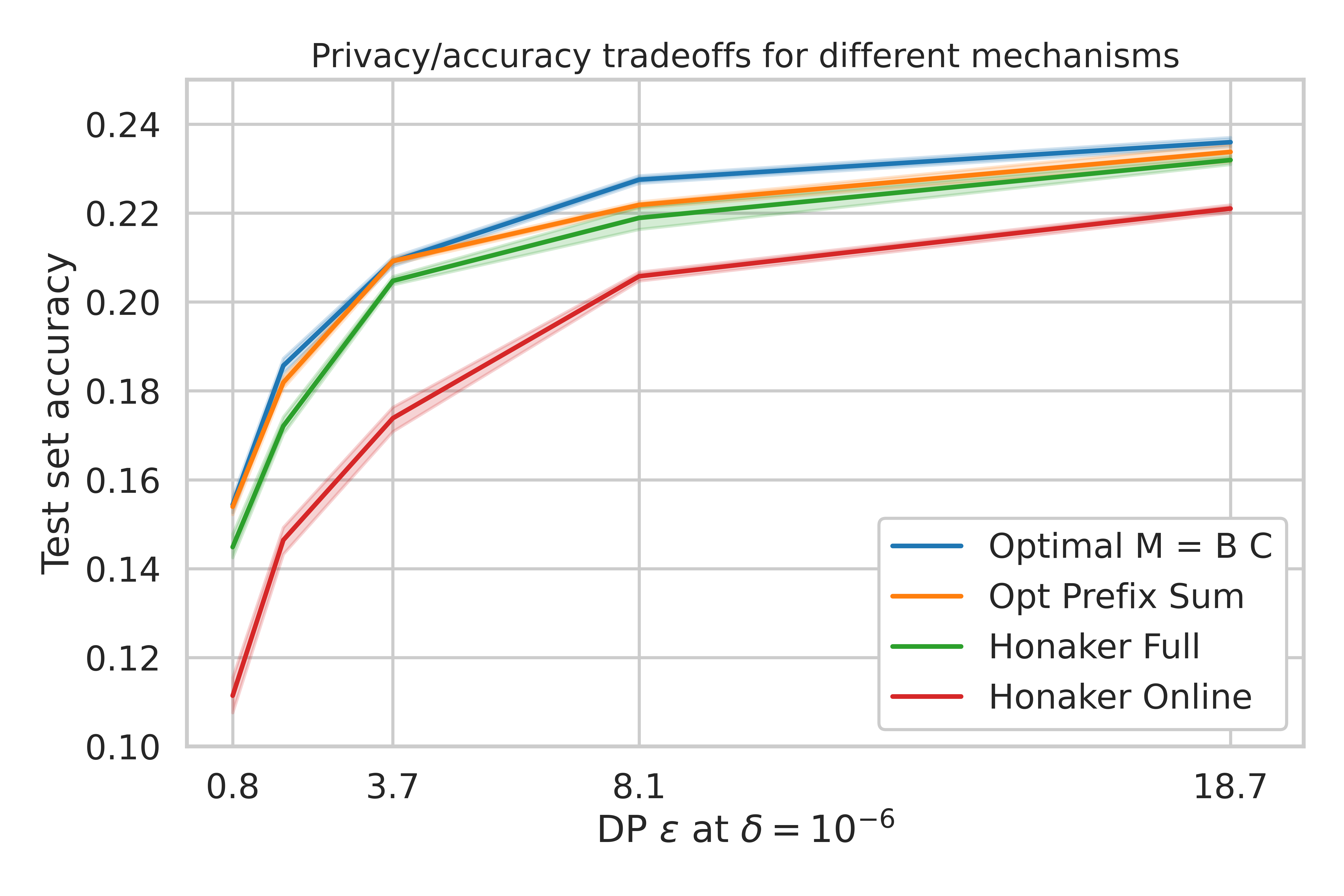}
    \vspace{-0.3in}
    \caption{Test accuracy %
    for the StackOverflow next-word-prediction task. A grid search over client and server learning rates and momentum $\beta$, with the best hyperparameters selected based on validation set accuracy. We then re-ran 11 repetitions with the best hyperparameters and report the mean test set accuracy with confidence intervals. All models trained with 100 clients per round and a constant learning rate schedule.}
    \label{fig:mechanism_test_error}
   \end{minipage}\hfill
    \begin{minipage}[t]{0.48\linewidth}
    \centering
    \includegraphics[width=1\linewidth]{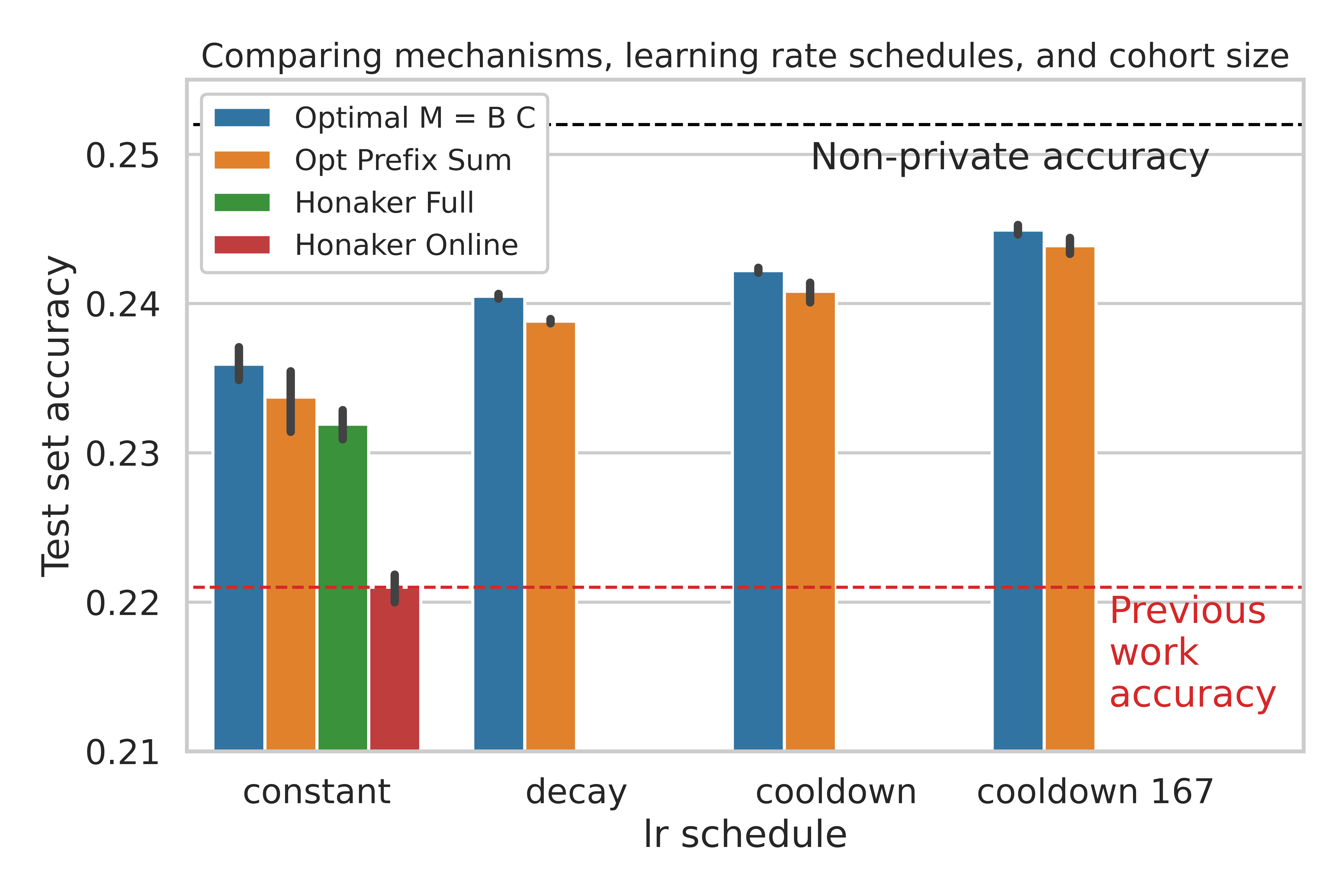}
    \vspace{-0.3in}
    \caption{Test accuracy for mechanisms incorporating learning rate decay in the manner of \cref{sec:more_mechanisms} at $\epsilon=18.9$. The red horizontal line represents test accuracy of previous state of the art at $\epsilon=18.9$; highest horizontal line, test accuracy of non-private model. The final bar group shows learning rate cooldown with $167$ clients/round, the maximum possible for a single training pass of 2048 training rounds.
    }
    \label{fig:lr_decay_test_acc}
    \end{minipage}
\end{figure}

\section{Experimental results}\label{sec:experiments}
The results of \cref{sec:streaming_priv,sec:more_mechanisms} significantly expand the space of mechanisms which can be used in training ML models with differential privacy in the single-pass setting. In this section we demonstrate that these techniques can in fact significantly advance the state-of-the-art in private ML. 

\mypar{User-level privacy for language models}
Private training is particularly important for generative language models: training language models on data from the right distribution is critical for utility (e.g., user input in a mobile keyboard \citep{NWP18,DPNWP}), but this data is often privacy-sensitive. Further, language models have been shown to be capable of memorizing training data \citep{carlini19secret,song2019auditing,carlini21extracting}. In this setting it is  important to consider user-level DP, where the neighbor relation of the DP guarantee covers all of the training examples (tokens) from any one user, as opposed to a single training example \citep{mcmahan2017learning}. In our setting, this corresponds to ensuring that each user's examples contribute to a bounded $\ell_2$-norm update to a single row of $\bfX$ (\cref{assumption:single_pass}). This is accomplished by extending \cref{alg:dpmfsgd} in the natural way to Federated Averaging \citep{FL1}: instead of a single gradient, we take $\obs_i$ to be the sum of the individually-clipped-to-$\clipnorm$ updates of all users (100 or 167 in our experiments) participating in the current round, with each user contributing to a single round over the course of training.

For these reasons, we focused on the StackOverflow next-word prediction problem, introduced in \citep{reddi20adaptive} and publicly hosted in TensorFlow-Federated (TFF) \citep{ingerman_ostrowski_2019}. This task was explored extensively in \citep{kairouz2021practical}, and serves as a major benchmark in federated learning, used in \citep{reddi20adaptive,kairouz2021practical,charles2021on,51174,fedrecon}, and others. The StackOverflow dataset contains sufficiently many clients to support single-pass algorithms with 100 clients per round, similar to the baseline setup of \citep{kairouz2021practical}. For this reason, we are able to provide true $(\epsilon, \delta)$ privacy quantifications for the models we train.\footnote{With 342,477 users this dataset is still small compared to many real-world applications (e.g., Gboard has 5 billion downloads). The extrapolations verified by \citep{mcmahan2017learning,kairouz2021practical} suggest the accuracy results of \cref{fig:lr_decay_test_acc} would hold with $(\eps=1.36, \delta=10^{-7})$-DP if the population and cohort size (clients per round) were both scaled by $10\times$.}

\mypar{Results} We compare the four mechanisms of \cref{tab:algs} on this problem; full experimental methodology, as well as additional plots (e.g., for validation error vs. training rounds) are provided in \cref{app:experiment_details}. \cref{fig:mechanism_test_error} shows that even with constant learning rates, our matrix factorization approaches significantly outperform the previous state-of-the-art across a range of privacy $\epsilon$'s. Further, thanks to the results of \cref{sec:streaming_priv}, we are able to apply the \HonakerFull mechanism for comparison.
\cref{fig:lr_decay_test_acc} shows that applying learning rate decay (dropping the learning rate by $0.15\times$ for the last 512 rounds) and learning rate cooldown (linearly dropping the learning rate from $1.0\times$ to $0.05\times$ over the last 512 rounds) show added improvements. These experiments all used 100 clients/round as in \citep{kairouz2021practical}, but the rightmost bars shows increasing this to 167 (the maximum possible for a single pass of 2048 rounds) provides additional accuracy. In combination, these techniques close more than 2/3rds of the gap between private and non-private training.

\section{Conclusions}\label{sec:conclustions}
We have shown the general applicability of the Gaussian matrix mechanism to the adaptive streaming setting, introduced a highly efficient mechanism of determining optimal (in the sense of total $\ell_2^2$ error) matrix mechanisms, used this approach to directly incorporate momentum and learning rate schedules into the DP mechanism, and empirically demonstrated the resulting private SGD (or FedAvg in the federated setting) substantially improves on the state of the art for private ML.

While our focus has been on the application of these techniques to gradient-based optimization algorithms, we emphasize that the problem of producing private estimates for linear queries in the adaptive streaming setting is a fundamental DP primitive of much broader applicability, as noted in the introduction, and so our work immediately leads to improvements in those applications as well.

Finally, our work raises numerous natural follow-up questions which we hope will inspire subsequent work; we sketch these in \cref{sec:future}.

\section*{Acknowledgements} We thank Zachary Charles, Thomas Steinke, Jonathan Ullman, Zheng Xu, and Anastasia Koloskova for their valuable feedback and insights. In particular Thomas pointed us to the Speyer’s argument of the lower bound; Zach discussed several of
the technical arguments with the authors; Jon provided helpful insights into alternate  proofs of  Theorem~\ref{thm:GaussianAdaptive}; Zheng provided valuable pointers to code which we were able to leverage; and Anastasia caught a significantly dropped term in~\cref{prop:sgd_utility}. 

Adam Smith was supported in part by NSF award CNS-2120667 and gifts from Apple and Google.

Sergey Denisov was supported by NSF award DMS-2054465 and Van Vleck Professorship research award.

\bibliographystyle{unsrtnat}
\bibliography{references}
\clearpage

\appendix

\section{Summary of notation and important matrices}\label{app:special_matrices}

The prefix-sum linear operator $\bfS$, and its inverse:

\begin{small}
\begin{equation}\label{eq:prefix_sum_s}
\bfS :=
\begin{pmatrix}
1 & 0 & 0 & \cdots & 0\\
1 & 1 & 0 & \cdots & 0\\
1 & 1 & 1 & \cdots & 0\\
\vdots & \vdots & \vdots & \ddots & \vdots \\
1 & 1 & 1 & \cdots & 1
\end{pmatrix}
\qquad
\text{and}
\qquad
\bfS^{-1} :=
\begin{pmatrix}
1 & 0 & 0 & \cdots & 0\\
-1 & 1 & 0 & \cdots & 0\\
0 & -1 & 1 & \cdots & 0\\
\vdots & \vdots & \vdots & \ddots & \vdots \\
 0& 0 & 0 & \cdots & 1
\end{pmatrix}.
\end{equation}
\end{small}

Representation of momentum SGD as a linear operator $\bfM = \ML \MR$:
\begin{small}
\begin{equation}\label{eq:momentum_matrices}
\ML :=
\begin{pmatrix}
\eta_1 & 0      & 0      & \cdots & 0\\
\eta_1 & \eta_2 & 0      & \cdots & 0\\
\eta_1 & \eta_2 & \eta_3 & \cdots & 0\\
\vdots & \vdots & \vdots & \ddots & \vdots \\
\eta_1 & \eta_2 & \eta_3 & \cdots & \eta_\dimension
\end{pmatrix}
\qquad
\text{and}
\qquad
\MR :=
\begin{pmatrix}
1       & 0      & 0    & \cdots & 0\\
\beta   & 1      & 0    & \cdots & 0\\
\beta^2 & \beta  & 1    & \cdots & 0\\
\vdots  & \vdots & \vdots & \ddots & \vdots \\
\beta^{\dimension-1} & \beta^{\dimension-2} & \beta^{\dimension-3} \cdots & 1
\end{pmatrix}
\end{equation}
\end{small}

\mypar{Summary of notation} 
The following table briefly summarizes notation used throughout this work.

\begin{table}[!h]
\renewcommand{\arraystretch}{1.4}%
\begin{tabular}{ll}

\hline
$\obs_i \in \R^\mdim$ & Input (e.g. gradient) on step $i$ of the online process. \\
$\obsM \in \R^\datadim$ & Matrix of all inputs, $\obs_i = \obsM\idx{i}{:}$. \\
$\bfA \in \R^{\dimension \times \dimension}$  & Lower-triangular linear query matrix to be factorized as $\bfA = \bfB \bfC$. \\
$\lambda_{\min}(\bfA)$, $\lambda_{\max}(\bfA)$. & Smallest and largest eigenvalues of real matrix $\bfA$. \\
$\bfA^*$  & Conjugate transpose of $\bfA$. \\
$\bfX^\star$  & A matrix $\bfX$ that is ``optimal'' in a context-dependent sense. \\
$\bfA^\dagger$  & Moore-Penrose pseudoinverse of matrix $\bfA$. \\
$\bfA\idx{i}{j}$ & The $(i, j)^{\text{th}}$ entry of matrix $\bfA$. \\
$\bfA\idx{i}{:}$ and $\bfA\idx{:}{j}$  & The $i^{\text{th}}$ row and $j^{\text{th}}$ column. \\
\hline
\end{tabular}
\end{table}

\section{Future work.}\label{sec:future}

Each of the sections above poses a unique set of problems for future investigation, many interrelated. We will highlight only some of the major questions left open by this work.

\mypar{Scalable mechanism implementations}

\cref{thm:GaussianAdaptive} shows that we need not restrict ourselves to any particular matrix structure in order to guarantee privacy over adaptive streams. \cref{sec:efficient} shows we can find efficient approximations for the case of prefix sums, but this leaves open the question of whether better or more general approximations are possible, or whether one can optimize over structures that allow efficient implementations directly. 

\mypar{Analysis and numerics of $\phi$}
 
\cref{thm:local_contraction} represents a usable convergence result for iterates of the mapping $\phi$; on the other hand, it represents only partial progress on the conjecture of global convergence of these iterates. Though we factorized many distinct matrices in the course of writing this paper, we generated no reason to doubt this conjecture. Indeed, the  speed of convergence of these iterates of $\phi$ (see \cref{app:numerics_of_phi}) only makes this method more intriguing from a theoretical perspective. Further, though the fixed-point method utilized to compute these factorizations has enabled significant exploration (as detailed in \cref{sec:more_mechanisms}), it still does not quite represent the optimal algorithm for computing these optima: an explicit formula for the fixed point of $\phi$ would clearly be desirable, and might yield interesting insights into the structure of these optimal matrices.

We finally note that for production use, additional care will be needed to ensure that claimed privacy guarantees fully account for floating point imprecision.

\mypar{Adaptive choice of the query}

While the sequence of gradients during optimization is adaptive (subsequent gradients depend on previous gradients), as we have seen SGD with momentum can be expressed as a fixed linear operator $\bfM$. Data-independent learning rate schedules can be incorporated into an optimization matrix in a similar fashion, again allowing for optimal DP matrix mechanisms. However, adaptive learning rate schedules such as AdaGrad amount to a \emph{non-linear} (and adaptive, not fixed) map on the gradient sequence; hence a very interesting open question is to see if the approach used here can be extended to adaptive optimization algorithms.

\section{Tree aggregation and decoding as matrix factorization} \label{app:tree_agg_matfac}
As mention in \cref{sec:intro}, the tree data structure $\calT$ is linear in the data matrix $\obsM$ (all of its internal nodes are linear combinations of the rows $\obsM$). Therefore the mapping $\obsM \to \calT$ can be represented as multiplication by a matrix.
We present a simple recursive construction of this matrix. The base case is the $1 \times 1$ matrix $[1]$, which we will denote by $\Ctree^{(1)}$; we will define $\Ctree^{(k)} \in \R^{(2^{k}-1) \times (2^{k-1})}$ to be the matrix constructed by duplicating $\Ctree^{(k-1)}$ on the diagonal, and adding one more row of constant $1$s. That is,
\begin{equation}\label{eq:binary_tree_h_def}
\Ctree^{(1)} :=
\begin{pmatrix}
1 
\end{pmatrix}, 
\Ctree^{(2)} :=
\begin{pmatrix}
1 & 0 \\
0 & 1 \\
1 & 1
\end{pmatrix}, 
\Ctree^{(3)} :=
\begin{pmatrix}
1 & 0 & 0 & 0\\
0 & 1 & 0 & 0\\
1 & 1 & 0 & 0 \\
0 & 0 & 1 & 0 \\
0 & 0 & 0 & 1 \\
0 & 0 & 1 & 1 \\
1 & 1 & 1 & 1 
\end{pmatrix}, 
\end{equation}
and so on. Each row of $\Ctree^{(k)} \obsM$ can be seen readily to correspond to a node of the binary tree $\calT$ constructed from $\obsM$, assuming $\dimension =  2 ^ {k-1}$ (possibly padding with zeros if needed).

With this construction, it is straightforward to represent both vanilla differentially-private binary tree aggregation and the Honaker variant as instantiations of the matrix factorization framework. For a vector $\bfx$ with $\dimension=2 ^ {k-1}$ entries, vanilla binary-tree aggregation can be represented as $\bfC = \Ctree^{(k)}$, $\bfB$ an appropriate $\{0, 1\}$-valued matrix satisfying $\bfB\bfC = \bfS$ for prefix-sum $\bfS$. The Honaker estimators can both be computed as (real-valued) matrices also satisfying $\bfB\bfC = \bfS$, and are in fact optimal: 

\begin{prop}\label{prop:honaker_optimal}
For the prefix-sum matrix $\bfS$ with $n=2 ^ {k-1}$ rows, the (non-streaming) Honaker fully efficient estimator represents the minimal-loss factorization for prefix sum $\bfS = \bfB\bfC$ for $\bfC = \Ctree^{(k)}$. This estimator is precisely $\bfS\bfC^\dagger$. The streaming Honaker estimator-from-below represents the minimal loss factorization satisfying the property that the $j^{th}$ row of $\bfB$ zeros out rows in the matrix $\bfC\obsM$ which place nonzero weight on the $i^{th}$ row of $\obsM$ for $i > j$. The Honaker estimator-from-below can be expressed similarly row-by-row with a constrained pseudoinverse of $\bfC$.
\end{prop}
\begin{proof}
We begin by recalling a geometric property of the Moore-Penrose pseudoinverse. Theorem 2.1.1 of \cite{nla.cat-vn1139431} states that for any matrix $\bfC \in \C^{m \times n}$, vector $\bfs \in \C^m$, the vector $\bfC^\dagger \bfs$ is the minimal least-squares solution to the linear system $\bfC\bfx = \bfs$. Notice that this statement is implicitly a statement of uniqueness; $\bfC^\dagger\bfs$ is the \emph{unique} minimal-norm solution to $\bfC\bfx = \bfs$, assuming feasability of this equation.
Since the square of the Frobenius norm of the matrix $\bfB$ is the sum of the squared norms of its rows, we may apply this Theorem row-by-row to $\bfB$ to demonstrate that the minimal Frobenius norm solution $\bfB$ to $\bfS = \bfB\bfC$ for fixed $\bfC$ is $\bfS\bfC^\dagger$.
 
This minimal Frobenius norm property may be translated to a statistical perspective. That is, for a fixed matrix $\bfC$ and data matrix $\obsM$, $\bfS\bfC^\dagger$ represents the minimal-variance unbiased linear estimator for $\bfS\obsM$ given the noisy estimates $\bfC\obsM + \bfZ$. This is precisely the definition of Honaker's fully efficient estimator in Section 3.4 of \cite{honaker_trick}, and we have the first statement of this proposition.

The second follows similarly, but leveraging instead the geometric properties of the constrained pseudoinverse. These properties are collected in Theorem 3.6.3 of \cite{nla.cat-vn1139431}, and allow us to compute directly the optimal $\bfB$ under constraints that certain entries in each row must be $0$, corresponding to the constraints stated in the proposition. By construction of the matrices $\Ctree^{(k)}$, the property described in the statement of \cref{prop:honaker_optimal} corresponds to restricting the linear estimator computed from a binary tree to depend only on the information below the nodes corresponding to the $1$s in a binary expansion of the index of the partial sum under consideration. This is precisely the definition of the estimator from below in Section 3.2 of \cite{honaker_trick}.
\end{proof}

\section{Proofs and missing details for Section~\ref{sec:streaming_priv}}
\label{app:streaming_priv}

\begin{proof}[Proof of Proposition~\ref{prop:adaptiveDPpure}]

The key idea is that the nonadaptive version of the definition implies a bound on the log-odds ratio that always holds (even after the fact). 

For simplicity, we focus on the case where the universe of possible outputs $\bfa$ is discrete (to avoid measurability issues). 

Fix an adversary $\adv$ and mechanism $\mech$. Recall $\side$ is fixed an unknown to the adversary. When $\side =0$, the probability of a particular view $(\obsM,\obsMp,\bfa)$ is the following. We write $(\obsM,\obsMp,\bfa) \gets \ip{\mech}{\adv}_0$ for the event with sequence of mechanism outputs $\bfa$, when the mechanism and the adversary are operating with the variable $\side=0$, and the neighboring data streams are $\obsM$ and $\obsMp$ (and analogously for $\side=1$). 

\begin{eqnarray*}
\Pr((\obsM,\obsMp,\bfa) \gets \ip{\mech}{\adv}_0) =&&\\ 
 \Pr\paren{ \adv() = (\obs_1,\obsp_1)} &\times& \Pr\paren{\mech(\obs_1)=\bfa_1} \times \\
 \Pr\paren{ \adv(\bfa_1) = (\obs_2,\obsp_2) \big| \obs_1,\obsp_1} &\times& \Pr\paren{\mech(\obs_2)=\bfa_2 \big| \obs_1,\bfa_1} \times \\
  &\cdots&  \\
 \underbrace{\Pr\paren{ \adv(\bfa_{\dimension-1}) = (\obs_\dimension,\obsp_\dimension) \big| \obs_1,...,\obs_{\dimension-1},\obsp_1,...,\obsp_{\dimension-1}}}_{\text{these do not depend on $\side$}} & \times & \underbrace{\Pr\paren{\mech(\obs_\dimension)=\bfa_\dimension \big| \obs_1,...,\obs_{\dimension-1},\bfa_1,...,\bfa_{\dimension-1}}}_{\text{these terms depend on $\side$}} \, .
\end{eqnarray*}
The probability of $(\obsM,\obsMp,\bfa)$ when $\side =1$ is similar, except that the inputs to $\mech$ are now $\obsp_t$'s instead of $\obs_t$'s. Either way, we get a product of $2\dimension$ terms, half of which are about the probability of $\adv$'s outputs, and half of which are about $\mech$'s outputs. They key fact here is that the terms describing $\adv$'s output are the same in both expressions. When we take the ratio, therefore, those terms cancel out and we obtain: 
\begin{multline*}
    \frac{\Pr((\obsM,\obsMp,\bfa) \gets \ip{\mech}{\adv}_0)}{\Pr((\obsM,\obsMp,\bfa) \gets \ip{\mech}{\adv}_1)} 
\\
= 
\frac
{
    \Pr\paren{\mech(\obs_1)=\bfa_1} \times \Pr\paren{\mech(\obs_2)=\bfa_2 \big| \obs_1,\bfa_1} \times \cdots \times \Pr\paren{\mech(\obs_\dimension)=\bfa_\dimension \big| \obs_1,...,\obs_{\dimension-1},\bfa_1,...,\bfa_{\dimension-1}}
}{
    \Pr\paren{\mech(\obsp_1)=\bfa_1} \times \Pr\paren{\mech(\obsp_2)=\bfa_2 \big| \obsp_1,\bfa_1} \times \cdots \times \Pr\paren{\mech(\obsp_\dimension)=\bfa_\dimension \big| \obsp_1,...,\obsp_{\dimension-1},\bfa_1,...,\bfa_{\dimension-1}}
}  
\\
= 
\frac{\Pr(\mech(\obs_1,...,\obs_\dimension) = (\bfa_1,...,\bfa_\dimension))}{\Pr(\mech(\obsp_1,...,\obsp_\dimension) = (\bfa_1,...,\bfa_\dimension))}
\, .
\end{multline*}
This last expression involves no adversary—it is simply the ratio of the probabilities that the mechanism would have produced a given output sequence if the sequences $\obsM$ and $\obsMp$ had been specified \textit{nonadaptively}.
Since $\obsM,\obsMp$ are always valid neighboring sequences, and since the nonadaptive mechanism's guarantee holds for all output sequences, the ratio above is bounded between $e^{-\eps}$ and $e^{\eps}$, as desired. 

\end{proof}

\begin{proof}[Proof of Theorem~\ref{thm:GaussianAdaptive}]
The idea is to show that, when $\bfA$ is lower triangular, the mechanism $\mech$ can be rewritten in such a way that the adaptive privacy of $\mech$ can be deduced from the privacy guarantees of the usual Gaussian mechanism with adaptively selected queries. 

Let $(\bfL,\bfQ)$ form a lower-triangular LQ-factorization of $\bfB$, meaning that $\bfL$ is lower triangular, $\bfQ$ is orthonormal, and $\bfB = \bfL\bfQ$. By assumption, $\bfA$ is square and invertible, so $\bfL$ and $\bfQ$ are also square and invertible.  Now consider the modified mechanism 
\[
\tilde\mech(\obsM) = \bfL(\bfQ\bfC \obsM + \noiseM)\quad \text{where}\ \noiseM\sim\calN(0,\kappa^2\sigma^2)^{\dimension\times d}
\]
where $\kappa\sigma$ is the same noise standard deviation as in the original mechanism. Since $\bfL$ and $\bfA$ are lower triangular, it also means that $\bfL^{-1}\bfA=\bfQ\bfC$ is also lower triangular. This means that $\bfQ\bfC\obsM + \noiseM$ can operate in the continuous release model, as row $i$ of $\bfQ\bfC\obsM$ depends only on the first $i$ rows of $\obsM$.

Next, we further show the mechanism $\bfQ\bfC \obsM + \noiseM$ (that is, $\tilde\mech$ without the post-processing operation of multiplying with $\bfL$) is an instance of the standard Gaussian mechanism for computing an adaptively defined function \emph{in the continuous release model} with a guaranteed bound on the global $\ell_2$ sensitivity.\footnote{Observe the same claim cannot be made for $\mech$, e.g., as $\bfC \obsM + \noiseM$ cannot be used in the continuous release setting as in general $\bfC$ induces a dependence on not-yet-seen data.}
Let $\obsM,\obsMp \in \calN$ be any two \textit{fixed} neighboring data streams with $\|\bfC(\obsM-\obsMp)\|_F\leq\kappa$. Then because $\bfQ$ is orthonormal we have $\|\bfQ \bfC(\obsM-\obsMp)\|_F\leq\kappa$. 
Letting $\bfg = \text{flatten}(\bfQ \bfC \obsM) \in \R^{\dimension \mdim}$ and $\bfh =  \text{flatten}(\bfQ \bfC \obsMp) \in \R^{\dimension \mdim}$, we have $\norm{\bfg - \bfh}{2} \le \kappa$. Hence, $\bfQ\bfC \obsM + \noiseM$ is equivalent to an application of the standard Guassian mechanism on inputs $\bfQ \bfC \obsM$, and the result for adaptive streams follows from Claim~\ref{cl:adapGauss} below. This claim holds as the privacy loss random variable is stochastically dominated by an appropriate normally-distributed random variable~(e.g., \cite{vadhan2017complexity}).

\begin{claim}[Folklore]
\label{cl:adapGauss}
Consider a streaming data vector $\bfg=[g_1,\ldots,g_n]\in\mathbb{R}^n$ s.t. for any neighboring stream $\bfh$ we have the $\ell_2$-sensitivity $\ltwo{\bfg-\bfh}\leq \kappa$. If $\bfg+\calN(0,\kappa^2\sigma^2)^n$ satisfy $(\epsilon,\delta)$-DP (or $\rho$-zCDP or $\mu$-Gaussian DP) in the \textit{nonadaptive} continual release model, then the same mechanism satisfies the same privacy guarantee in the adaptive continuous release model.
\end{claim}

As $\bfL$ is lower-triangular, the adaptive streaming DP guarantee of $\bfQ\bfC \obsM + \noiseM$ extends to the mechanism $\tilde\mech$ by the post-processing property of DP.

Finally, we have
\[
\mech(\obsM) %
             = \bfB(\bfC \obsM + \noiseM)
             = \bfL(\bfQ\bfC \obsM + \bfQ \noiseM),
\]
and so the only difference from $\tilde\mech$ is the use of noise $\bfQ \noiseM$ vs $\noiseM$. Since $\bfQ$ is orthonormal and the Gaussian distribution is rotationally invariant, $\bfQ \noiseM$ and $\noiseM$ are identically distributed, and hence $\mech$ and $\tilde\mech$ produce identical output distributions on any fixed data set $\obsM$. Thus an adversary can simulate $\mech$ given access to $\tilde\mech$. This in turn means that the privacy guarantee of the mechanism $\tilde\mech$ transfers to the mechanism $\calM$.
This completes the proof. 
\end{proof}

\begin{proof}[Proof of \cref{prop:lt_factorizations}]
The existence of such a lower-triangular factorization with identical induced matrix mechanism distribution follows directly from the proof of \cref{thm:GaussianAdaptive}. The body of the proof leverages the distributional equivalence of all mechanisms expressible as

$$\left(\bfB\bfU\right)\left(\bfU^*\bfC\right).$$
Since $\bfA$ is lower-triangular, letting $\bfU = \bfR^*$ recovers a lower-triangular mechanism (IE, both terms in the factorization are lower-triangular) which is distributionally equivalent to the factorization $\bfA = \bfB\bfC$.

The claimed uniqueness follows from uniquness of the QR factorization with all-nonnegative diagonal entries; see, e.g., \citep[Theorem 2.1.14]{horn_and_johnson}.
\end{proof}

\subsection{Not all additive noise mechanisms are adaptively private}
\label{sec:nonadaptivePrivNonGaussian}

Consider the following two step sampling process\footnote{For the clarity of noation, in this section we will refer to random variables with uppercase, and their corresponding instantiations with lower case. Additionally, all norms are $\ltwo{\cdot}$.}:
\begin{enumerate}
    \item $A\sim \calN(0,\id_d)$.
    \item $B \sim \calN(0,\Sigma)$ where $\Sigma = \id -  (1-\eta)\frac{AA^t}{\norm{A}{}^2}$ and $\eta = 5\frac{\sqrt{\log(d)}}{d}$. Observe that, conditioned on $A=a$, we can write $B$ as the sum of independent random variables $B=B_1+B_2$ where $B_1 \sim \calN(0,\id - \frac{aa^t}{\norm{a}{}^2})$---a component orthogonal to $a$ with variance 1 in all other directions--- and $B_2 \sim \eta \frac{a}{\norm{a}{}} \cdot \calN(0,1)$---a component parallel to $a$ with much smaller variance $\eta$ in that direction.
    \item Return $(A,B)$
\end{enumerate}

Now consider a mechanism $\mech$ that takes a parameter $\sigma$ and two inputs of the form  $(x_1,x_2) \in (\R^d)^2$ where $x_1=0$ (always) and $x_2$ has Euclidean norm at most 1, and returns $(x_1 + \sigma A, x_2 +\sigma B)$.  The mechanism can be run interactively, in which case $x_2$ could be selected based on $A$, which can be deduced from $x_1 + \sigma A$. The notion of neighboring here is trivial: every pair $(0,x_2)$ is a neighbor of every other pair $(0,y_2)$ so long as $\norm{x_2}{}$ and $\norm{y_2}{}$ are at most 1. 

For simplicity, we formulate the mechanism for the special case when $n=2$ and the first input is forced to be 0, but similar constructions and reasoning apply for larger $n$ and other types of inputs.

\begin{prop}
    $\mech$ is nonadaptively $(\eps,\delta)$-DP with parameters  $\eps =\Theta( \sqrt{\ln(1/\delta)}/\sigma)$ when $d$ is sufficiently large and $\delta \geq \exp(-cd)$ for an absolute constant $c>0$.
\end{prop}

\begin{proof}
    Let $(0,x_2)$ and $(0,y_2)$ be the inputs submitted by the adversary. Let $W = \ip{x_2-y_2}{\tfrac{A}{\norm{A}{}}}$. Observe that $\ip{x_2-y_2}{A}$ distributed as $N(0,\norm{x_2-y_2}{}^2)$ (with variance at most 2) and that $\norm{A}{}$ is between $\frac 1 2 \sqrt{d}$ and $2\sqrt{d}$ with probability $1-\exp(-\Omega(d))$ by standard concentration arguments.  Thus, $W$ is at most $\eta = \frac{5\sqrt{\log(d)}}{d}$ with probability $1-\exp(-\Omega(d))$.
    
    Given $A=a$, we can write the output $b=b_1+b_2$ as a sum of a component $b_1$ parallel to $a$ and a component $b_2$ orthogonal to $a$. Recalling the notation $B=B_1+B_2$ from the definition of $(A,B)$, we get the following distributions when $\side=0$: 
    \begin{align*}
        b_2 = & \paren{\ip{\tfrac{a}{\norm{a}{}}}{x_2} + \eta\sigma Z}\tfrac{a}{\norm{a}{}} \text{ where }Z\sim N(0,1) \, \text{ and} \\
        b_1 = &\Pi(x_2)+B_2\text{ where $\Pi$ is the projector onto the subspace orthogonal to $a$. }
    \end{align*}
    We get the same distribution with $\side=1$, except that $x_2$ is replaced by $y_2$. Conditioned on $a$, we have additive noise with a well-understood distribution in both cases. The likelihood ratio thus depends only on $W$ and $\Pi(x_2-y_2)$. 
    
    The first component consists of adding noise with standard deviation $\eta\sigma$ to an input with sensitivity $|W|\leq \frac{5\sqrt{\log(d)}}{d}$; the second consists of adding noise in with standard deviation $\sigma$ (in all $d-1$ relevant dimensions) to an input with sensitivity at most 2. Each of these satisfy $(\eps,\delta)$-differential privacy for $\eps= \Theta(\sqrt{\ln(1/\delta)}/\sigma)$, as desired.
\end{proof}

\begin{prop} When $\sigma\eta<1/3$, the mechanism
    $\mech$ is not adaptively $(\eps,\frac 1 4)$-DP unless $\eps \geq \frac{1}{3(\sigma \eta)^2}$.
\end{prop}

\begin{proof} An adaptive adversary first submits vectors $x_1, y_1$ (both 0), receives a first output $a$ which is either $x_1+A$ or $y_1+A$, and then submits $x_2$ and $y_2$ and receives a second output $b$ which is either $x_2+B$ or $y_2+B$. 
Consider the specific adversary submits $x_2 = \frac{a}{\norm{a}{}}$ and $y_2=-x_2$ (based on the first output $a$) and then receives output $b$. 

The idea is that the variance of $B$ in the direction of $x_2=a$ is only $\eta\sigma$ (instead of $\sigma$) and so---informally---the effective $\epsilon$ of the mechanism is roughly $1/(\eta\sigma)$ instead of $1/\sigma$. When $d$ is large, $\eta$ is much smaller than 1 and so the mechanism provides much weaker privacy guarantees in the adaptive setting.

More formally,  consider the random variable $\ip{a}{b}$. 
The component of $B$ in the direction of $a$ can be written $B_2 = \eta \frac{a}{\norm{a}{}} \cdot Z$ for $Z\sim N(0,1)$. When $\side=0$, we thus have
\[
\ip{a}{b} = \ip{a}{x_2 +\sigma B} = \ip{a}{x_2 +\sigma B_2} =\ip{a}{\tfrac{a}{\norm{a}{}} + \sigma \eta \tfrac{a}{\norm{a}{}} Z} = \norm{a}{} (1 + \sigma \eta Z)\, .
\]
Similarly, when $\side =1$, the inner product $\ip{a}{b}$ is distributed as $\norm{a}{} (-1 + \sigma \eta Z)$. The probability that $\ip{a}{b}>0$ is at least $\frac 1 2$ when $\side =1$ and, for $\sigma\eta<1$, the same probability is at most $\exp\paren{-\frac{1}{2(\sigma \eta)^2}}$ when $\side =0$. In particular, the mechanism is not $(\eps,\delta)$-DP in the adaptive model unless $\frac 1 2 \leq e^\eps \Pr(\ip{a}{b}>0 | \side =0) -\delta$; that is, it requires $\eps \geq \frac{1}{2(\sigma \eta)^2} - \ln(\frac{2}{1-2\delta})$. The bound is at least $\frac{1}{3(\sigma \eta)^2}$ for $\delta \leq 1/4$ and $\sigma \eta < 1/3$. 
\end{proof}

\section{Proofs and observations for \cref{sec:computing_optima}}
\label{app:iterated_convergence}

\subsection{Proof of Theorem~\ref{thm:matmech_privacy_quant}}
\label{app:gaussM}

\begin{proof}
The proof essentially follows from standard arguments about the DP guarantee for the Gaussian mechanism~\cite{ODO,mironov2017renyi}. In the following, we provide some of the details for completeness.

First, notice that it is sufficient to state that the computation $\bfC\obsM+\noiseM$ satisfies $(\epsilon,\delta)$-DP, due to the post processing property of DP. Now consider two data sets $\obsM$ and $\obsMp$ differing in one data record (as per the neighborhood definition in~\ref{def:diiffP}). We have $\bfC(\obsM-\obsMp)$ is equal to the outer product $\bfc\obs$, where $\obs$ is the row of $\obsM$ that was changed, and $\bfc$ is the corresponding column of $\bfC$. By assumption in the theorem statement, we have 
\[
\left\|\bfc\bfg\right\|_F\leq \ltwo{\bfc}\cdot\ltwo{\bfg}\leq\gamma\zeta.
\]
With the bound on the sensitivity above, if each entry of $\bfZ$ is drawn i.i.d. from $\calN(0,\sigma^2)$, then $\bfC\obsM+\bfZ$ satisfies $\rho=\frac{\gamma^2\zeta^2}{\sigma^2}$-zCDP (Definiton~\ref{def:zCDP})~\cite{bun2016concentrated}. We set the noise standard deviation $\sigma$ by the use of Remark 15 in~\citet{steinke_composition}. Correspondingly, we have $(\epsilon,\delta)$-DP~\cite{dwork2014algorithmic}.
\end{proof}

\subsection{Proof of \cref{thm:expressions_for_x}}
\begin{proof}
For simplicity we consider the equality-constrained version of \cref{eq:symmetrized_problem} (permissible by \citep{opt_convex_fact}): 
\begin{equation}\label{eq:symmetrized_problem_eq}
    \xopt = \argmin_{\bfX \text{ is PD}, \bfX\idx{i}{i} = 1, i \in [\dimension]} \tr(\bfA^* \bfA \bfX^{-1}).
\end{equation}
We begin by noting that Slater's condition (see \citep[Section 5.2.3]{boyd_convex_opt}) holds in our setting, since the minimum eigenvalue of a matrix is a concave function (expressible as a minimum of linear functions), and we know from \citep{opt_convex_fact} that that the optimum is strictly positive definite. Therefore strong duality holds, and complementary slackness implies we may drop the positive-definiteness constraint when we move to the Lagrange formulation.
Thus, we introduce a Lagrange multiplier $\dualv$ for \cref{eq:symmetrized_problem_eq}, defining,

\begin{align}
  L(\bfX, \dualv) 
  &= \tr(\bfA^*\bfA\bfX^{-1}) + \sum_{i=1}^\dimension \dualv_i (\bfX_{i, i} - 1) \notag \\
  &= \tr(\bfA^*\bfA\bfX^{-1}) + \tr\big(\diag(\dualv) (\bfX - \bfI)\big). \label{eq:lagrangian_def}
\end{align}

Differentiating \cref{eq:lagrangian_def} with respect to $\bfX$, we find
\begin{equation}\label{eq:lagrangian_grad}
  \frac{\partial L}{\partial \bfX} = -(\bfX^{-1}\bfA^* \bfA\bfX^{-1}) + \diag\left(\dualv\right).
\end{equation}

Let $\xopt$ be the optimizer of the primal problem \cref{eq:symmetrized_problem_eq}; then, by the Lagrange multiplier theorem (see, e.g., Proposition 4.1.1 of \citep{Bertsekas99}), there exists a unique $\vopt$ satisfying
\begin{equation}\label{eq:lagrangian_zero_def_necessary}
   \diag\left(\vopt\right) = {\xopt}\inv \bfA^* \bfA{\xopt}\inv,
\end{equation}
which is an equivalent form of \cref{eq:lagrangian_zero_def}. Since $\bfA$ is full-rank, and $\xopt$ is known from \citep{opt_convex_fact} to be positive-definite, \cref{eq:lagrangian_zero_def_necessary} implies that $\diag\left(\dualv\right)$ is invertible (indeed, positive definite).

Solving \cref{eq:lagrangian_zero_def_necessary} for $\bfX$ corresponds to solving for a generalized matrix square root. The equation \cref{eq:lagrangian_zero_def_necessary} may be uniquely solved, yielding
\begin{equation}
  \Xofv\left(\dualv\right)= \diag(\dualv)^{-1/2} \left(\diag(\dualv)^{1/2}\bfA^*\bfA\diag(\dualv)^{1/2}\right)^{1/2} \diag(\dualv)^{-1/2}. \tag{\ref{eq:x_for_dualv}}
\end{equation}
Clearly the $\Xofv(\vopt)$ defined by \cref{eq:x_for_dualv} represents a solution for \cref{eq:lagrangian_zero_def_necessary}; that $\Xofv(\vopt) = \xopt$  can be seen by substituting \cref{eq:lagrangian_zero_def_necessary} for $\diag(\vopt)$ in \cref{eq:x_for_dualv}, and evaluating the result to the form $ \xopt$.

Since $\xopt$ has constant 1s on the diagonal (by the formulation \cref{eq:symmetrized_problem_eq}), the expression \cref{eq:x_for_dualv} implies that 
\[
\diagpart\left(\sqrt{\diag(\vopt)^{1/2} \bfA^*\bfA \diag(\vopt)^{1/2}}\right) = \vopt,
\]
and that therefore $\vopt$ is a fixed point of the mapping $\phi$ defined by \cref{eq:phi_def}.

We have shown that an optimizer $\xopt$ corresponds to a fixed point $\vopt$ of $\phi$. If we begin with a fixed point $\vopt$ of $\phi$, and define $\Xofv(\vopt)$ via \cref{eq:x_for_dualv}, the preceding calculations show that $\Xofv(\vopt)$ is both feasible and a stationary point of the Lagrangian. The strict convexity of the problem, along with its smoothness, imply that the Hessian of the Lagrangian is positive definite at this stationary point, and therefore (e.g. by Proposition 4.2.1 of \citep{Bertsekas99}), this $\Xofv(\vopt)$ is a local minimizer. Strict convexity implies that this local minimizer is in fact the global minimizer.

The final claim of \cref{eq:lower_bound_from_dual} follows immediately from weak duality and the fact that
\[
\inf_{\bfX} L(\bfX, \dualv) = L(\Xofv(\dualv), \dualv))
\]
since the problem on the left is convex in $\bfX$, and hence \cref{eq:x_for_dualv} gives an optimality condition. \cref{eq:lower_bound_from_dual} follows by using $\bfA^* \bfA \bfX\inv = \bfX\diag\left(\vopt\right)$ in the first trace in \cref{eq:lagrangian_def}, and then simplifying using properties of the matrix trace.
\end{proof}

\subsection{Proof of local-contractive property of $\phi$.}\label{app:proof_of_phi_convergence}

Recall that we study the map, defined in \cref{eq:phi_def},
\[
\phi(\bfv) := \diagpart\left(\sqrt{\diag(\bfv)^{1/2} \,\bfB^*\bfB\,\diag(\bfv)^{1/2}}\right),
\]
 from the positive cone in $\R^\dimension$ to itself. By \cref{thm:expressions_for_x}, we know that it has a unique fixed point, which we denote by $\vopt$. We will need some notation:\smallskip
 
 \begin{itemize}
 \item $\bfQ=\sqrt{\bfB^*\bfB}$.
\item In $\R^\dimension$, we consider two inner products
 \[
 \langle \bfx, \bfy\rangle=\sum_{j}x_jy_j, \quad \langle \bfx, \bfy\rangle_1=\sum_{j}x_jy_jw_j, \quad w_j^{-1}=v^*_j\,.
 \]
The fist one is Euclidean, the second one is weighted with the weight given by $\vopt$ itself. The corresponding norms are $\|\bfx\|$ and $\|\bfx\|_1$.
\item The operator norms of linear map $\bfA$ acting in $\R^\dimension$ will be denoted 
\[
\|\bfA\|, \,\, \|\bfA\|_1
\]
depending on the considered inner products, e.g.,
\[
\|\bfA\|_1=\sup_{\bfx: \|\bfx\|_1=1}\|\bfA \bfx\|_1\,.
\]
\end{itemize}

We start by giving a simple estimate on $\vopt$.
\begin{prop}\label{p0}Suppose $\bfQ$ satisfies
\begin{equation}\label{h1}
0<\kappa_1\le \bfQ\le \kappa_2
\end{equation}
with some constants $\kappa_1$ and $\kappa_2$.
Then, 
\[
\kappa_1^2\le \diag \vopt\le \kappa_2^2\,.
\]
\end{prop}
\begin{proof}
Given any two non-negative matrices $\bfA$ and $\bfB$ that satisfy $\bfA\le \bfB$, we clearly have
\begin{equation}\label{a1}
\diagpart \bfA\le \diagpart \bfB, \quad \sqrt \bfA\le \sqrt \bfB\,.
\end{equation}
Then, 
\[
\kappa_1^2\le \bfQ^2\le \kappa_2^2 \Rightarrow \kappa_1^2(\diag \vopt)\le (\diag \vopt)^{1/2}\bfQ^2(\diag \vopt)^{1/2}\le \kappa_2^2(\diag \vopt)\,.
\]
Thus, we apply \eqref{a1} by first taking the square roots and then the diagonal parts of both sides to get 
\[
\kappa_1\sqrt{\diag \vopt}\le \diag \vopt\le \kappa_2\sqrt{\diag \vopt}
\]
after we recall that $\vopt$ is the fixed point of $\phi(\bfv)$. The required statement is now immediate.
\end{proof}
\noindent {\bf Remark.} The argument in the proof shows that $\phi$ maps the convex set $\{\bfv: \alpha\le \diag \bfv\le \beta\}$ into itself provided that $0<\alpha\le C_1$ and $C_2\le \beta$. Since $\phi$ is continuous, the Brouwer fixed point theorem gives yet another proof that a fixed point of $\phi$ exists.\bigskip

The map $\bfv \mapsto \phi(\bfv)$ is smooth on $\R^\dimension$. Its derivative at point $\vopt$ is therefore a linear map in $\R^\dimension$. We will denote

\begin{equation}\label{eq:l_def}
\bfL:=D\phi(\vopt).
\end{equation}

Our central result is precisely the statement that the (weighted) norm of $\bfL$ is smaller than 1, and hence $\phi$ is a local contraction around $\vopt$:

\begin{thm}\label{sd4}The map $L$ is a contraction in weighted norm, i.e., 
\[
\|\bfL\|_1\le C(\kappa_1,\kappa_2)<1\,.
\]
\end{thm}

\noindent {\bf Remark.} This immediately implies that the sequence $\{\bfv_n\}$ given by $\bfv_{n+1}=\phi(\bfv_n)$ converges exponentially fast to $\vopt$ when $\bfv_0$ is chosen sufficiently close to $\vopt$.
The exact parameters here depend only on $\kappa_1$ and $\kappa_2$. The size of the neighborhood in which the first-order approximation implies that $\phi$ itself is a contraction similarly depends on $\kappa_1$, as this controls the smoothness of $\phi$.\bigskip

We will recall some facts before giving the proof of this theorem.

\begin{prop}\label{p1}
If $\bfA$ is $\dimension\times \dimension$ matrix and $\bf d \in \R^{\dimension}$, then
\begin{multline}\label{a2}
\left(\bfA+\frac{1}{2} (\diag \bfd ) \bfA+\frac{1}{2}\bfA(\diag \bfd )\right)^2=\\\bfA^2(\diag \bf d ) +(\diag \bf d )\bfA^2+\bfA(\diag  d ) \bfA+O(\| d \|^2)
\end{multline}
where the constants in $O$ depend on $\|\bfA\|$ only.
\end{prop}
\begin{proof}That is an immediate calculation.
\end{proof}
\begin{prop}\label{p2}If $\bfA$ is $\dimension\times \dimension$ positive matrix, then
\begin{equation}
\sqrt \bfA=\frac{\bfA}{\pi}\int_0^\infty (\bfA+t)^{-1}\frac{dt}{\sqrt t}\,.
\end{equation}
\end{prop}
\begin{proof}
That follows from the Spectral Theorem for symmetric matrices and the trigonometric integral formula
\[
\sqrt \lambda=\frac{\lambda}{\pi}\int_0^\infty (\lambda+t)^{-1}\frac{dt}{\sqrt t}, \quad \lambda>0\,,
\]

which follows by substituting $t = \tan^2\theta$.
\end{proof}

\begin{prop}
If $\bfA, \bfV$ are $\dimension\times \dimension$ matrices and both $\bfA$ and $\bfA+\bfV$ are non-degenerate, then
\[
(\bfA+\bfV)^{-1}=\bfA^{-1}-(\bfA+\bfV)^{-1}\bfV\bfA^{-1} \quad
\]
and 
\begin{equation}\label{a3}
(\bfA+\bfV)^{-1}=\bfA^{-1}-\bfA^{-1}\bfV\bfA^{-1}+O(\|\bfV\|^2). 
\end{equation}
\end{prop}
\begin{proof}To check the first identity, it is enough to multiply it from the left by $\bfA+\bfV$ and from the right by $\bfA$. The second identity will follows by iterating the first identity once.
\end{proof}

The formula for $\bfL$ is given in the following lemma.
\begin{lem}If $\bfT:=\sqrt{(\diag \vopt)^{1/2}\bfQ^2(\diag \vopt)^{1/2}}$, then
\[
\bfL \bfw=\bfw-\pi^{-1}\diagpart \left( \int_0^\infty (\bfT^2+t)^{-1}\bfT (\diag\bfw)(\diag \vopt)^{-1}\bfT (\bfT^2+t)^{-1}\sqrt tdt         \right)\,.
\]
\end{lem}
\begin{proof}
This result is based on a long but straightforward calculation. First, introduce $\bfd \in \R^\dimension$ by
\[
\diag \bfv=(\diag \vopt) \exp(\diag \bf d )\,.
\]
Hence, denoting $\bf\Delta:=\diag \bf d $ for shorthand, one has
\[
\phi(\bfv)=\diagpart\left(\sqrt{\bfT^2+0.5{\bf\Delta} \bfT^2+0.5\bfT^2 \bf\Delta}\right)+O(\|{\bf\Delta}\|^2)
\]
since $\bfT>0$ and the matrix square-root is Lipschitz-continuous at every point which represents a positive matrix. If one denotes
$\bfX:=\sqrt{\bfT^2+0.5{\bf\Delta} \bfT^2+0.5\bfT^2 \bf\Delta}$, then  \cref{p1,p2} yield
\[
\bfX-(\bfT+0.5{\bf\Delta} \bfT+0.5\bf\Delta\bfT)=
\]
\[
\frac{\bfX^2}{\pi}\int_0^\infty(\bfX^2+t)^{-1}\frac{dt}{\sqrt t}-\frac{\bfX^2+\bfT{\bf\Delta}\bfT}{\pi}\int_0^\infty(\bfX^2+\bfT{\bf\Delta}\bfT+t)^{-1}\frac{dt}{\sqrt t}=
\]
\[
\stackrel{\eqref{a3}}{=}\frac{\bfT^2}{\pi}\int_0^\infty(\bfT^2+t)^{-1}\bfT{\bf\Delta}\bfT(\bfT^2+t)^{-1}\frac{dt}{\sqrt t}-\bfT{\bf\Delta}+O(\|{\bf\Delta}\|^2).
\]
If we recall that $ \vopt=\phi(\vopt)=\diagpart \bfT$, then
\begin{multline}
\diagpart \bfX=\diagpart (\bfT+0.5{\bf\Delta} \bfT-0.5 \bfT {\bf\Delta})+\\\diagpart \frac{\bfT^2}{\pi}\int_0^\infty(\bfT^2+t)^{-1}\bfT{\bf\Delta}\bfT(\bfT^2+t)^{-1}\frac{dt}{\sqrt t}+O(\|{\bf\Delta}\|^2)=\\
\diagpart \bfT+\diagpart \frac{1}{\pi}\int_0^\infty (\bfT^2+t-t)(\bfT^2+t)^{-1}\bfT{\bf\Delta}\bfT(\bfT^2+t)^{-1}\frac{dt}{\sqrt t}+O(\|{\bf\Delta}\|^2)=\\
\phi(\vopt)+\diagpart (\bfT {\bf\Delta})-\pi^{-1}\diagpart \left( \int_0^\infty (\bfT^2+t)^{-1}\bfT{\bf\Delta}\bfT (\bfT^2+t)^{-1}\sqrt tdt         \right)+O(\|{\bf\Delta}\|^2)=\\
\phi(\vopt)+\diagpart((\diag \vopt) {\bf\Delta})-\pi^{-1}\diagpart \left( \int_0^\infty (\bfT^2+t)^{-1}\bfT{\bf\Delta}\bfT (\bfT^2+t)^{-1}\sqrt tdt         \right)+O(\|{\bf\Delta}\|^2)
\end{multline}
Now, notice that 
\[
\diag\bfv=\diag\vopt+\diag \vopt{\bf\Delta}+O(\|\bf\Delta\|^2)
\]
and
\[
{\bf\Delta}=(\diag \bfw)(\diag \vopt)^{-1}+O(\|\bf\Delta\|^2)\,
\]
where $\bfw:=\bfv-\vopt$. Finally, we have
\begin{multline}
\phi(\bfv)=\phi(\vopt)+\bfw-\\\frac{1}{\pi}\diagpart \left( \int_0^\infty (\bfT^2+t)^{-1}\bfT  (\diag{\bfw})(\diag \vopt)^{-1} \bfT (\bfT^2+t)^{-1}\sqrt tdt         \right)+O(\|{\bfw}\|^2)
\end{multline}
and that proves the required statement.
\end{proof}
Our next step is to obtain the matrix representation of $\bfL$ in the standard basis of $\R^\dimension$.
\begin{lem}If $(\bfT^2+t)^{-1}\bfT=:\bfC(t)=\bfC_{[i,j]}(t)$, then
\[
\bfL=\bfI-\Bigl\{  \frac{1}{\pi}\int_0^\infty (\bfC_{[i,j]}(t))^2\frac{\sqrt t}{v^*_j}dt\Bigr\}, \quad i,j\in \{1,\ldots,\dimension\}\,.
\]
\end{lem}
\begin{proof} That calculation is straightforward after we use  symmetry of matrices $\bfT$ and $\bfC$. \end{proof}

The Theorem \ref{sd4} will be proved if \cref{lem:estimate_of_l} is shown. In its proof, the following 
property of the Schur (elementwise, also known as Hadamard) product of two matrices is used.
\begin{prop}Suppose $\bfA$ and $\bfB$ are non-negative matrices of size $\dimension\times \dimension$. Then,
\[
\lambda_{\min}(\bfA\circ \bfB)\ge \lambda_{\min}(\bfA)\lambda_{\min}(\bfB)\,.
\]
\end{prop}
\begin{proof}Indeed, the matrix $\bfA\circ \bfB$ represents the principal submatrix of the Kronecker (or tensor) product $\bfA \otimes \bfB$. Since $\lambda_{\min}(\bfA \otimes \bfB)=\lambda_{\min}(\bfA)\lambda_{\min}(\bfB)$, we get our result.
\end{proof}

\begin{lem}\label{lem:estimate_of_l}
The operator $\bfL$ is selfadjoint with respect to the weighted inner product and $\|\bfL\|_1\le C(\kappa_1,\kappa_2)<1$.
\end{lem}
\begin{proof}
First, we can write a bilinear form
\[
\langle \bfL \bfv, \bfw\rangle_1=\langle  \bfv, \bfw\rangle_1-\frac{1}{\pi}\int_0^\infty \sum_{i,j=1}^\dimension\bfG_{[i,j]}(t)\frac{v_jw_i}{v^*_jv^*_i}\sqrt tdt
\]
where $\bfG:=\bfC\circ \bfC$, the Schur product, is a symmetric matrix. Hence, $\langle \bfL \bfv, \bfw\rangle_1=\langle \bfL \bfw, \bfv\rangle_1$ and therefore $\bfL$ is appropriately selfadjoint. Next, we will prove a bound
\begin{equation}\label{schur_estimate}
 C(\kappa_1,\kappa_2)\|\bfv\|_1^2\le \frac{1}{\pi}\int_0^\infty \sum_{i,j=1}^\dimension\bfG_{[i,j]}(t)\frac{v_jv_i}{v^*_jv^*_i}\sqrt tdt\le C_5\|\bfv\|_1^2
\end{equation}
with some positive $C$ and $C_5\in (0,1)$. That estimate for quadratic form is sufficient to prove that $\|\bfL\|_1<1$ due to the variational characterization of the norm of a self-adjoint operator, i.e., $\|\bfL\|_1=\sup_{\bfv: \|\bfv\|_1=1}|\langle \bfL\bfv, \bfv\rangle_1|$.

We claim that 
\begin{equation}\label{h3}
\int_0^\infty \bfG(t)\sqrt t dt\ge C_3(\kappa_1,\kappa_2)>0
\end{equation}
in a sense of positive matrices. Indeed, 
\[
\lambda_{\min} (\bfG(t))\ge (\lambda_{\min}(\bfC(t)))^2
\]
as follows from the properties of the Schur product. Since
$
\bfC(t)={\bfT}/{(\bfT^2+t)}
$, we get
\[
\int_0^\infty (\lambda_{\min}(\bfC(t)))^2  \sqrt tdt \ge C_3(\kappa_1,\kappa_2)>0
\]
where $C_3$ depends on parameters $\kappa_1$ and $\kappa_2$ from \eqref{h1} only. So, our claim \eqref{h3} is proved. Given \eqref{h3}, we can write
\begin{eqnarray*}
\int_0^\infty \sum_{i,j=1}^\dimension \bfG_{[i,j]}(t)\frac{v_jv_i}{v^*_jv^*_i}\sqrt tdt\ge C_3(\kappa_1,\kappa_2)\sum_{j=1}^\dimension \left|\frac{v_j}{v^*_j}\right|^2\ge\\ C_4(\kappa_1,\kappa_2)\sum_{j=1}^\dimension \frac{|v_j|^2}{|v^*_j|}=C_4(\kappa_1,\kappa_2)\|\bfv\|_1^2
\end{eqnarray*}
thanks to Proposition \ref{p0}. This shows the left bound in \cref{schur_estimate}.

The inequality 
\[
 \frac 1\pi\sum_{i,j=1}^\dimension\bfG_{[i,j]}(t)\frac{v_jv_i}{v^*_jv^*_i}\sqrt tdt\le C_5\|\bfv\|_1^2
\]
is equivalent to 
\begin{equation}\label{p9}
 \frac 1\pi \sum_{i,j=1}^\dimension\bfG_{[i,j]}(t)\frac{x_jx_i}{\sqrt{v^*_jv^*_i}}\sqrt tdt\le C_5\|\bfx\|^2
\end{equation}
if we make the change of variables $x_j:=v_j/\sqrt{v^*_j}, j\in \{1,\ldots,\dimension\}$. It will be convenient to introduce a symmetric matrix $\bfD$ with coefficients given by 
\[
\bfD_{[i,j]}=\frac 1\pi\int_0^\infty \bfG_{[i,j]}(t)\frac{1}{\sqrt{v^*_jv^*_i}}\sqrt tdt=\frac 1\pi\int_0^\infty (\bfC_{[i,j]}(t))^2\frac{1}{\sqrt{v^*_jv^*_i}}\sqrt tdt\,.
\]
To bound the norm of this matrix, we will start with the following observation. The application of Spectral Theorem to matrix $\bfT$ yields
\[
\frac{1}{\pi}\int_0^\infty (\bfT^2+t)^{-2}\bfT^2\sqrt tdt=C_5\bfT
\]
where
\[
C_5=\frac{1}{\pi}\int_0^\infty (1+\xi)^{-2}\sqrt \xi d\xi=\frac{2}{\pi}\int_0^\infty (1+u^2)^{-2}u^2du<\frac{2}{\pi}\int_0^\infty (1+u^2)^{-1}du=1\,.
\]
Recall also that $\vopt=\diagpart \bfT$ and therefore
\[
\frac{1}{\pi} \int_0^\infty \diagpart((\bfT^2+t)^{-2}\bfT^2)\sqrt tdt=C_5\vopt
\]
Since the matrix elements of $\bfC(t)=(\bfT^2+t)^{-1}\bfT$ are given by $\bfC_{[i,j]}(t)$, the diagonal elements of $(\bfT^2+t)^{-2}\bfT^2$ can be obtained by the formula
\[
\sum_{j=1}^\dimension \bfC_{[i,j]}(t)\bfC_{[j,i]}(t)=\sum_{j=1}^\dimension (\bfC_{[i,j]}(t))^2
\]
for $i\in \{1,\ldots, \dimension\}$. Therefore, we get an identity
\[
\frac{1}{\pi}\int_0^\infty \sum_{j=1}^\dimension (\bfC_{[i,j]}(t))^2\sqrt tdt=C_5v^*_i, \quad i\in \{1,\ldots, \dimension\}
\]
which can be rewritten as 
\[
\sum_{j=1}^\dimension \bfD_{[i,j]}\sqrt{v_i^*v_j^*}=C_5v^*_i, \quad i\in \{1,\ldots,\dimension\}\,.
\]
The elements $\bfD_{[i,j]}$ are non-negative and $\bfD_{[i,j]}=\bfD_{[j,i]}$. Taking the vector $\{\sqrt{v^*_i}\}$ with positive entries, we rewrite the previous identity as
\[
\sum_{j=1}^\dimension \bfD_{[i,j]}\sqrt{v_j^*}=C_5\sqrt{v^*_i}, \quad i\in \{1,\ldots,\dimension\}\,.
\]
The application of Schur's test for the norm of matrix gives $\|\bfD\|\le C_5<1$. Since $\bfD$ is symmetric, this bound implies \eqref{p9}.

\end{proof}

\subsection{Numerical observations of the map $\phi$.}\label{app:numerics_of_phi}

Some care must be taken with floating point issues in the implementation of the map $\phi$. In particular, numerical evaluation of $\phi$ depends critically on the computation of a matrix square root, and precision in this computation is crucial for the usability of these fixed-point methods. 

Several numerical approaches can improve the stability of these algorithms. In particular, some of the expressions above (e.g., the definition of $\Xofv$) imply  a priori lower bounds on the eigenvalues of matrices for which we need square roots. The results of \citep{merikoski_kumar} yield straightforward lower bounds that can stabilize our iterative algorithms. These bounds can be applied to ensure the iterates never encounter pathological numerical artifacts. As the size of matrices scales up (in particular, our factorizations usually focused on $2048 \times 2048$ matrices, and larger matrices are of interest), we observed that performing all computations in \texttt{float64} precision was crucial to minimizing these numerical artifacts.

We observed experimentally that while factorizing some matrices, though the fixed-point method itself converged independently of the matrix factorized, some oscillation in the values of the loss \cref{eq:l_expression} occurred. Further investigation is needed to determine whether this oscillation represents a true feature of the iterated dynamics, or simply another numerical artifact, due e.g. to lack of precision in the matrix square root. If the former, certain approaches to prove global convergence of these iterates are ruled out: in particular, those which rely on this loss as a potential function, which iterating $\phi$ always decreases.

\begin{figure}[t]
    \centering
    \makebox[\textwidth][c]{\includegraphics[width=0.8\textwidth]{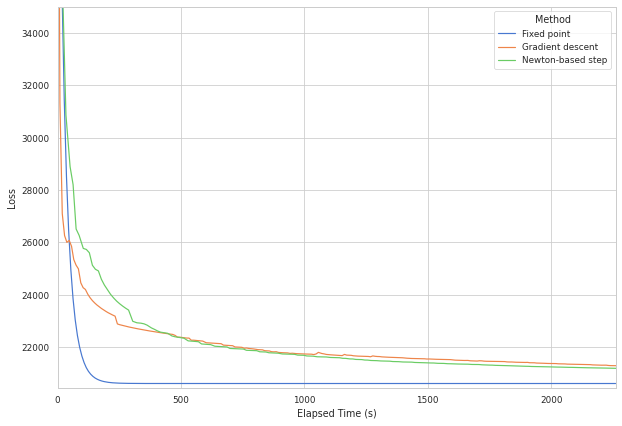}}%
    \caption{Value of loss \cref{eq:l_expression} against elapsed time for a gradient-descent based, Newton-direction based, and fixed-point implementation of computing optimal factorizations of 2048-dimensional prefix sum matrix $\bfS$. The gradient descent and Newton direction-based methods used an Armijo step size search, and checked for existence of Cholesky factorization to verify positive-definiteness of the iterates, as suggested by \citep{opt_convex_fact}. The methods were initialized identically, leveraging the expression \cref{eq:x_for_dualv}, a significantly better initialization for the gradient-based methods than might be obvious in the absence of this expression (e.g., initialization to $\bfI$).}
    \label{fig:convergence_speed}
\end{figure}

To evaluate empirical usefulness of this fixed-point method, we implemented three different algorithms for computing optimal factorizations:
\begin{enumerate}
    \item A gradient-descent-based procedure to compute the optima of \cref{eq:symmetrized_problem}, guaranteed to be convergent by the convexity of the problem.
    \item\citep[Algorithm 1]{opt_convex_fact}, a Newton-direction-based algorithm with global convergence guarantees, hand-optimized with the structure of the problem---in particular, avoiding the need to materialize a Hessian with $n^4$ elements. This implementation used the default settings from \citep{opt_convex_fact}.
    \item Simply iterating the mapping $\phi$.
\end{enumerate}

In all situations we tested, the fixed-point method was significantly faster than either of the other two, up to two orders of magnitude in some cases. In \cref{fig:convergence_speed}, we plot loss against time for an example of $2048\times2048$ matrix factorization using CPUs. The methods are all similarly amenable to GPU acceleration.

\clearpage
\section{Proofs for~\cref{sec:more_mechanisms}}

\subsection{Proof of~\cref{prop:sgd_utility}}

\begin{proof}
Following the analysis of Theorem C.1,~\citet{kairouz2021practical}, we introduce a hypothetical `unnoised' model trajectory $\widetilde{\theta_t}$. Define $b_t := \theta_t - \widetilde{\theta_t}$, and note that $b_t = -\eta \bfB_{[t, :]}\bfZ$.

We note the well-known equivalence of FTRL and gradient descent, with requilarization parameter $\lambda$ equivalent to $\frac{1}{\eta}$ (as can be seen by solving for the FTRL update). Following the standard linearization method of online convex optimization, we see:

\begin{equation*}
    \frac{1}{\dimension}\sum_{t=1}^\dimension\ell(\theta_t;\chi_t)-\ell(\thetaopt;\chi_t) \leq \frac{1}{\dimension}\sum_{t=1}^\dimension \ip{\nabla_t}{\theta_t - \thetaopt}
    = \underbrace{\frac{1}{\dimension}\sum_{t=1}^\dimension \ip{\nabla_t}{\widetilde{\theta_t} - \thetaopt}}_\text{\clap{~A}} + \underbrace{\frac{1}{\dimension}\sum_{t=1}^\dimension \ip{\nabla_t}{ \theta_t - \widetilde{\theta_t}}}_\text{\clap{~B}}
\end{equation*}

Similarly to~\citet{kairouz2021practical}, the term A may be handled with standard online convex optimization techniques, yielding
\begin{equation*}
    \text{\clap{A}}~\leq \eta L^2 + \frac{1}{2 \eta \dimension}\left(\|\thetaopt\|_2^2 - \|\theta_1\|_2^2\right),
\end{equation*}

so we are left to evaluate the expectation of B over the noise injected by~\cref{alg:dpmfsgd}. We compute:

\begin{align*}
    \mathbb{E}\left[\frac{1}{\dimension}\sum_{t=1}^\dimension \ip{\nabla_t}{ \theta_t - \widetilde{\theta_t}}\right] &\leq \frac{1}{\dimension} \mathbb{E} \left[ \sum_{t=1}^\dimension \|\nabla_t\|_2 \|\theta_t - \widetilde{\theta_t}\|_2 \right] \quad \quad &\text{Cauchy-Schwartz}\\
    &\leq \frac{L}{\dimension}\mathbb{E}\left[\sum_{t=1}^\dimension \|b_t\|_2\right]\\
    & \leq L \mathbb{E}\left[ \left(\frac{1}{\dimension} \sum_{t=1}^\dimension \|b_t\|_2^2\right)^{1/2}\right] &\text{Jensen's inequality}\\
    &\leq \frac{L}{\sqrt{\dimension}} \left(\mathbb{E} \left[\sum_{t=1}^\dimension \|b_t\|_2^2\right]\right)^{1/2} &\text{Jensen again}\\
    &=\frac{L\eta}{\sqrt{\dimension}} \left(\mathbb{E}\left[\|\bfB\bfZ\|_F^2\right]\right)^{1/2} &\text{definition of } b_t\\
    &=\frac{L\sigma\eta}{\sqrt{\dimension}}\|\bfB\|_F &\text{evaluating the expectation.}
\end{align*}

Putting together the estimates of A and B yields the result.

\end{proof}

\clearpage
\section{Visualization of Optimal Factorizations}\label{sec:optfactfig}
\newcommand{\LRA}{\bfL} %
\newcommand{\LRB}{\bfR} %
\newcommand{\Bopt}{\bfB^\star}
\newcommand{\Copt}{\bfC^\star}
\newcommand{\approxB}{\hat{\bfB}}
\newcommand{\lowrank}{r}
\newcommand{\ndiags}{h}
\newcommand{\buff}{\boldsymbol\beta}
\newcommand{\mask}{\bfU}

\begin{figure}[H]
    \centering
    \includegraphics[width=5in]{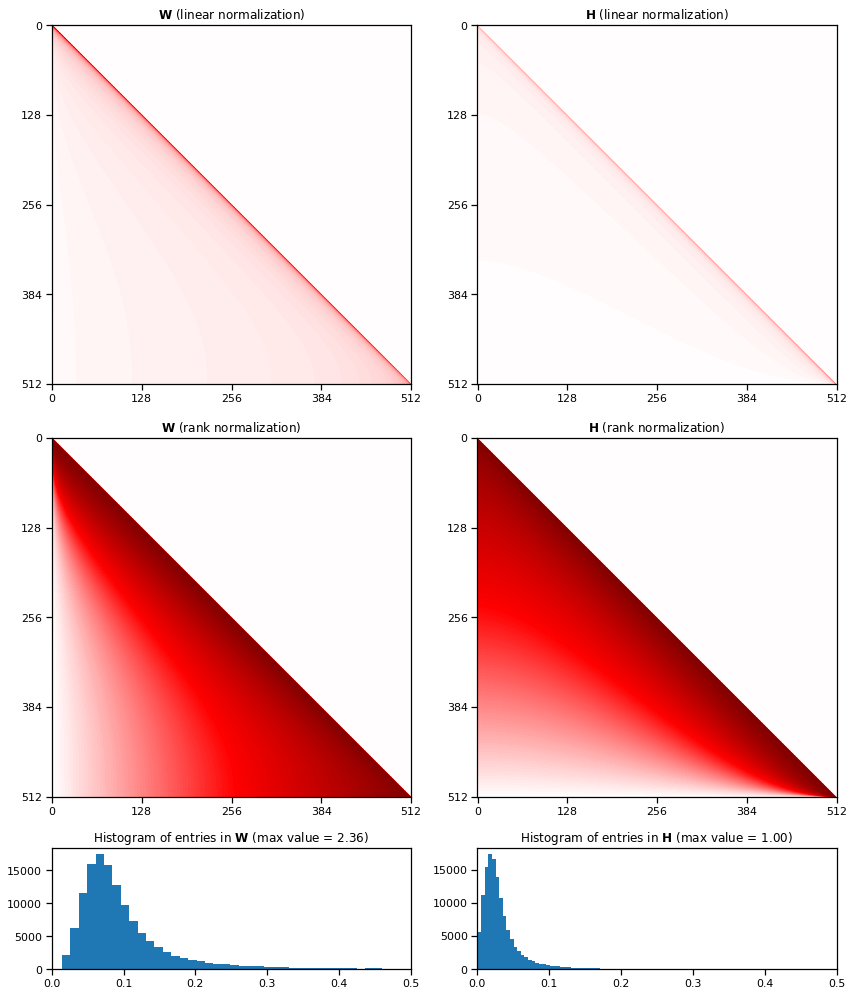}
    \caption{%
    Visualizations of the optimal streaming matrix factorization $\bfS = \bfW \bfH$ $(\bfB=\bfW, \bfC=\bfH)$ for cumulative sums with $\dimension=512$. 
    The matrix visualizations use a color palette that maps scalars in $[0, 1]$ to colors from white to dark red.
    The first row normalizes entries to $[0, 1]$ by simply dividing all entries by 2.36 (the largest value in either matrix). This clearly shows the heavy diagonal in $\bfB$.
    The second row normalizes the values in each matrix by ranking them by magnitude, and then mapping the ranks to $[0, 1]$ so 0 entries (the smallest) are mapped to 0.0, the median value is mapped to 0.5 (mid-red), and the largest value is mapped to 1.0 (darkest red). This visualization more clearly shows the off-diagonal structure. 
    The final row gives a histogram of the magnitudes of the non-zero entries in each matrix.}
    \label{fig:opt_w_h_vis}
\end{figure}

\begin{table}[t]
    \centering
    \renewcommand{\arraystretch}{1.3}
    \begin{tabular}{cS[table-format=3.1]S[table-format=3.1]S[table-format=3.1]c}
         \textbf{$\dimension$} 
           & \textbf{Honaker} 
           & \textbf{$(\Bopt, \Copt)$} 
           & \textbf{Efficient}
           & \textbf{$(\ndiags, \lowrank)$}\\
         \hline
         $2^8 = 256$     &  74.4 &  40.4 &  40.4 & (4, 4) \\
         $2^9 = 512$     & 116.5 &  62.0 &  62.2 & (5,4)\\
         $2^{10} = 1024$ & 180.8 &  94.6 &  95.5 & (5, 5)\\
         $2^{11} = 2048$ & 278.3 & 143.6 & 145.8 & (6, 5)\\
         $2^{12} = 4096$ & 425.6 & 217.3 & 224.0 & (6, 6)\\
         \hline
    \end{tabular}
    \vspace{0.1in}
    \caption{Values of $\sqrt{\calL}$ for the expected squared reconstruction error $\calL$ defined in \cref{eq:l_expression} (which implies equivalent levels of privacy). The ``Efficient'' column gives $\sqrt{\calL}$ for the structured approximation $\approxB$ of $\Bopt$ with parameters $(\ndiags, \lowrank)$ described below. When $\dimension=2^i$ we choose $\ndiags + \lowrank = i$, so that the mechanism based on $\approxB$ has memory and computation efficiency comparable to the Honaker approach.}
    \label{tab:loss_values}
\end{table}

\section{Computational efficiency for the matrix mechanism}\label{sec:efficient}
Our primary goal has been to develop mechanisms with best-possible privacy vs utility tradeoffs in the streaming setting. However, the $(\bfB, \bfC)$ we compute are in general dense, and do not obviously admit a computationally-efficient implementation of the associated DP mechanism. In contrast, tree aggregation (including, with a careful implementation, the streaming Honaker estimator) allows implementations with only $\log(\dimension)$ overhead; that is, each DP estimate of the $i^{th}$ partial sum can be computed in time and space $\calO(\mdim \log(\dimension))$.

In this section, we demonstrate empirically that the optimal $\Bopt$ from the factorization of the prefix sum matrix $\bfS$ can be approximated by structured matrices in such a way as to be competitive with the tree-aggregation approach in terms of computation and memory, but retain the advantage of substantially improved utility. Recalling \cref{alg:dpmfsgd}, the key is to compute $\bfB\idx{i}{:} \bfZ$ efficently. If $\bfB$ is arbitrary, this takes $\calO(\dimension \mdim)$ operations, which is likely prohibitive.

However, having a structured matrix $\approxB$ that allows efficient multiplication with $\bfZ$ mitigates this problem. We propose the following construction, which empirically provides a good approximation while also allowing computational efficiency. Let $\bfD^{(\ndiags)}$ denote the lower-triangular banded matrix formed by taking the first $\ndiags$ diagonals of $\bfB$, so $\bfD^{(0)}$ is the all-zero matrix, $\bfD^{(1)}$ is the main diagonal of $\bfB$, and $\bfD^{(2)}$ contains the main diagonal and one below it, etc.  Let $\mask^{(\ndiags)} \in \{0, 1\}^{n \times n}$ contain a $1$ in the place of each non-zero element of $\bfB$ not captured in $\bfD^{(\ndiags)}$ and zero elsewhere, so in particular $\bfB = \bfB \odot \mask^{(\ndiags)} + \bfD^{(\ndiags)}$ where $\odot$ is elementwise multiplication. Then, we propose the representation
\[
\approxB = \big(\LRA \LRB^\top\big)\odot \mask^{(\ndiags)} + \bfD^{(\ndiags)},
\]
where $\LRA, \LRB \in \R^{\dimension \times \lowrank}$. Finding a low-rank factorization $\LRA \LRB^\top$ which minimizes $\|\approxB - \bfB\|_F^2$ can be cast as a matrix completion problem, as we only care about approximating with $\LRA \LRB^\top$ the entries of $\bfB$ selected by $\mask^{(\ndiags)}$. For these experiments we used an alternating least squares solver with a regularization penalty of $10^{-6}$ on $\|\LRA\|^2_F + \|\LRB\|^2_F$ \citep{srebro04max,koren09matrix,jain2012lowrank}.
Given such a representation, the cost of computing $\approxB\idx{i}{:} \bfZ$ is $\calO((\ndiags + \lowrank) \mdim)$: we maintain accumulators $\buff$ such that 
\[
(\LRA \buff)\idx{i}{:} = \big((\LRA \LRB\tp  \odot \mask^{(\ndiags)}) \bfZ\big)\idx{i}{:},
\]
and $\buff$ can be updated in time $\lowrank \mdim$ on each step. Then, Finally, $(\bfD^{(\ndiags)}\bfZ)\idx{i}{:}$ can be computed in time $\ndiags \mdim$. \cref{alg:efficient} makes this algorithm explicit.

Columns 3 and 4 in \cref{tab:loss_values} shows empirically that this approximation recovers almost all of the accuracy improvement of $(\Bopt, \Copt)$ at comparable computational efficiency to tree aggregation with the Honaker estimator (that is, we choose $\ndiags + \lowrank = \log_2(n)$). While a paired $\bfC$ is not used directly in computing the private estimates, it is necessary in order to compute the loss $\calL$ defined in \cref{eq:l_expression}, as well as to appropriately calibrate the noise to achieve a DP guarantee (see \cref{thm:matmech_privacy_quant}). For these purposes an optimal $\bfC_{\approxB}$ can be found analogous to \cref{eq:constrained_problem} as $\bfC_{\approxB} = \approxB\inv \bfS$. 

\algrenewcommand{\algorithmiccomment}[1]{\hskip3em\# #1}

\begin{algorithm}[H]
\caption{An efficient implementation (executed by the trusted curator)}
\label{alg:efficient}
\begin{algorithmic}[1]
\State \text{\# Iterations and matrices/vectors are zero indexed (unlike elsewhere)}
\State Parameters:
\State \myindent Matrix $\bfD^{(\ndiags)}$ containing $\ndiags \in \{0, \dots, n\}$ diagonals from $\bfB$
\State \myindent Matrices $\LRA, \LRB \in \R^{n \times \lowrank}$
\State \myindent Noise matrix $\bfZ \in \R^\datadim$
\State \myindent Observations $\obsM \in \R^\datadim$
\State $\buff \assign  \mathbf{0} \in \R^{\lowrank \times \mdim}$ \Comment{Buffer for relevant part of $\LRB\tp \bfZ$.}
\State $\bfs \assign 0 \in \R^\mdim$ \Comment{Accumulator for prefix sum}
\For{$i$ in $1, \dots, n$} 
    \State $\bfs \plusequal \obsM\idx{i}{:}$  \Comment{Maintain the un-noised cumulative sum}
    \State $\result \assign 0 \in \R^\mdim$  \Comment{Accumulator for total noise in $i$th prefix sum}
    \For{$k$ in $0, \dots, \min(i, \ndiags-1)$} \Comment{Handle $\ndiags$ diagonals directly; No-op if $\ndiags=0$}
      \State $\result \plusequal \bfD^{(\ndiags)}\idx{i}{i-k} \bfZ\idx{i-k}{:}$  \Comment{$\ndiags \mdim$ multiplies}
     \EndFor
    \If{$i \ge \ndiags$} \Comment{Compute the low-rank portion}
      \State $i' \leftarrow i - \ndiags$
      \State $\buff \plusequal \LRB\tp\idx{i'}{:} \bfZ\idx{i'}{:} $ \Comment{$\lowrank \mdim$ multiplies}
      \State $\result \plusequal \LRA\idx{i}{:} \buff$  \Comment{$\lowrank\mdim$ multiplies}
    \EndIf
    \State Release $\bfs + \result$ \Comment{A DP estimate of $\sum_{t=1}^i \obsM\idx{t}{:}$}
\EndFor
\end{algorithmic}
\end{algorithm}

\section{Experiment Details}\label{app:experiment_details}
\mypar{Mechanism implementation}
Though \cref{sec:efficient} shows that time- and space-bounded approximations to our optimal factorizations are possible, for our experimental results we followed \cref{alg:dpmfsgd} and implemented the straightforward version of our mechanism. That is, we leverage the expression 

\[
\bfB\left(\bfC\obsM + \bfZ\right) = \bfA\obsM + \bfB\bfZ,
\]

for $\bfA = \bfB\bfC$, where $\bfA$ represents the linear operator we are interested in estimating. By introducing a seed to the generation of the noise vector $\bfZ$, the appropriate noise vector $(\bfB\bfZ)\idx{i}{:}$ can simply be computed afresh for each iteration of training (or round in the federated setting). The linear operators $\bfA$ in which we are interested admit efficient implementations; e.g., gradient descent with momentum can be implemented with a single buffer, representing the current state of the model. The computation of $\bfB\noiseM$ is therefore the dominant component in the above.

We normalized all of our factorizations to have sensitivity exactly 1 in the single-pass setting.

\mypar{Integration with federated learning} We implemented these mechanisms via the \texttt{DPQuery} interface in TensorFlow-Privacy \citep{TFpriv}, which integrates naturally with \texttt{tff.aggregators}, the aggregators library of TensorFlow-Federated. We were therefore able to reuse precisely the same code for training as \citep{thakkar2019differentially}, simply swapping in our matrix-factorization-based aggregators as an argument to TFF's \texttt{tff.learning.build\_federated\_averaging\_process} function. In conjunction with this paper, we are in the process of open-sourcing the code to reproduce our experiments. TFF's distributed C++ runtime, equipped with one machine for every 10 clients per round and low-priority CPU resources, enabled our experiment grids (including evaluation) to finish in approximately 1 day.

\mypar{Stackoverflow settings}

The preprocessing of our data, in addition to model architecture as well as the settings of various task-specific hyperparameters like the maximum number of examples processed per-client, we share with \citep{kairouz2021practical}.

\mypar{Test accuracy details} Test accuracies (excluding predictions on out-of-vocabulary and end-of-sentence tokens)  plotted against $\epsilon$ values associated to $\delta = 10^{-6}$ for various instantiations of the mechanisms we tested can be found in \cref{fig:mechanism_test_error}. This figure was generated with a sweep over client and server learning rates, with grids chosen via in the heatmap for FedAvgM in Figure 2 of \citep{reddi20adaptive}, as well as a sweep over server momentum values.
The noise multiplier settings were chosen with a simple calculation, based on the reported noise multipliers for StackOverflow NWP in \citep{kairouz2021practical}. In particular, by explicitly calculating the sensitivity of the binary tree (as in Theorem 4.1 of \citep{kairouz2021practical}), one can normalize the noise multipliers to be equivalent between the two settings. In the process of testing our code, we verified that we observed similar results to those claimed there under this normalization. The smallest noise multiplier in our setting corresponds to the largest $\epsilon$ in figure 2(a) of \citep{kairouz2021practical}, though our plots are not exactly comparable to theirs due to the different number of rounds in the two experimental setups. We calculate our $\epsilon$ values by simply measuring the privacy cost of the appropriate high-dimensional Gaussian query, by \cref{thm:dpmfsgd}. The grid we swept over can be found in \cref{tab:grids}.

\begin{figure}[!h]
    \centering
    \includegraphics[width=0.7\linewidth]{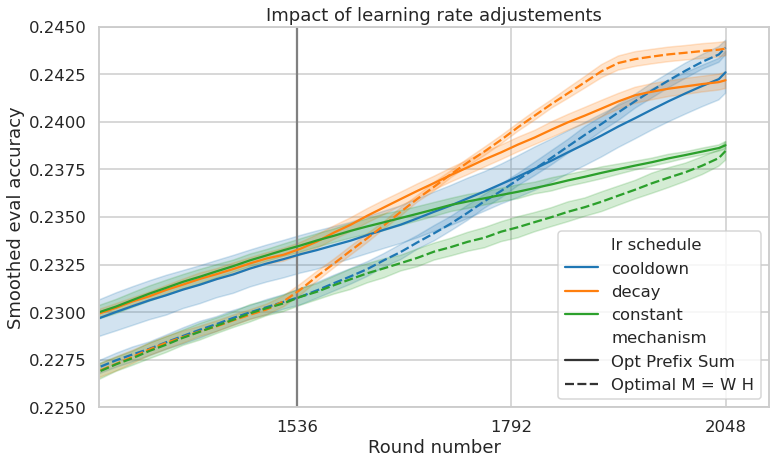}
    \caption{Smoothed validation accuracy over the final 748 rounds at $\epsilon=18.9, \delta=10^{-6}$, comparing momentum and learning rate decay implemented as postprocessing operations to the prefix-sum mechanism versus capturing these operations in the mechanism itself (see \cref{sec:more_mechanisms}). Vertical line represents the start of the decay schedule.}
    \label{fig:lr_mom_eval}
    \end{figure}

The error bars in \cref{fig:mechanism_test_error,fig:lr_decay_test_acc} were generated by first filtering down to runs which did not diverge from repeated runs with 10 seeds (at least 7 converged for each setting in \cref{fig:mechanism_test_error}, at least 8 for each setting in \cref{fig:lr_decay_test_acc}), then computing the empirical standard deviation. A similar process was used for \cref{fig:lr_mom_eval}.

\mypar{Evaluation accuracy details}

During training, we monitored performance on an evaluation set consisting of 10,000 sentences from outside of the training and test sets. We plot this evaluation accuracy for the final portion of training our learning-rate schedule and momentum matrices in \cref{fig:lr_mom_eval}.
\mypar{Additional figures}
We note that \cref{fig:lr_mom_eval} demonstrates a consistent artifact we witnessed in training these models: the momentum matrix factorization performs worse  than the prefix-sum matrix during the body of training, but catches up and overtakes towards the end of the training procedure. We hypothesize this to be an artifact of the way these mechanisms distribute variance on the operator residuals, and consider it an interesting pointer for future mechanism design, while noting that it implies the matrix factorizations are significantly tuned to the number of iterations for which they are designed.

\mypar{Hyperparameter settings} The parameter settings for the various figures in the main body and appendix can be found below.

\begin{table}[ht]
\caption{Grids used in initial search for \cref{fig:mechanism_test_error}.}
\centering
\begin{tabular}{l l} 
 \hline
 \textbf{Parameter} & \textbf{Grid values}\\ [0.5ex] 
 \hline
 Client learning rate & [0.5, 1.0] \\ 
 Server learning rate & [0.25, 0.5, 1.0, 2.0] \\ 
 Server momentum & [0.0, 0.9, 0.95] \\ 
 Noise multiplier & [0.341, 0.682, 1.364, 2.728, 5.456]\\ [1ex]
 \hline
\end{tabular}
\label{tab:grids}
\end{table}

\begin{table}[ht]
\caption{Hyperparameter settings for \cref{fig:mechanism_test_error}.}
\centering
\renewcommand{\arraystretch}{1.2}
\begin{tabular}{|c|c|c|} 
 \hline
 \textbf{Mechanism} & $\epsilon$
  & \textbf{(Server LR, Client LR, Server momentum)}\\ [0.5ex] 
 \hline
 \multirow{5}{8em}{\HonakerFull} & 18.9
  & $(0.5, 1., 0.95)$ \\
  & 8.2 & $(0.25, 1., 0.95)$ \\

  & 3.7 & $(0.25, 1., 0.9)$ \\
  
  & 1.7 & $(0.25, 0.5, 0.9)$ \\
  & 0.8 & $(0.25, 0.5, 0.0)$ \\
  \hline
  \multirow{5}{4em}{\HonakerOnline} & 18.9
  & $(0.25, 1., 0.95)$ \\
  & 8.2 & $(0.25, 1., 0.9)$ \\
  & 3.7 & $(0.25, 0.5, 0.9)$ \\
  & 1.7 & $(0.25, 1., 0.0)$ \\
  & 0.8 & $(0.25, 0.5, 0.0)$ \\
  \hline
  \multirow{5}{4em}{\OptPrefixSum} & 18.9
  & $(0.5, 1., 0.95)$ \\
  & 8.2 & $(0.25, 0.5, 0.95)$ \\
  & 3.7 & $(0.25, 1., 0.9)$ \\
  & 1.7 & $(0.25, 0.5, 0.9)$ \\
  & 0.8 & $(0.5, 0.5, 0.0)$ \\
  \hline
  \multirow{5}{5em}{\OptMomentum} & 18.9
  & $(1., 1., 0.9)$ \\
  & 8.2 & $(0.25, 1., 0.95)$ \\
  & 3.7 & $(0.25, 0.5, 0.9)$ \\
  & 1.7 & $(0.25, 0.5, 0.9)$ \\
  & 0.8 & $(0.5, 0.5, 0.0)$ \\
  [1ex]
 \hline
\end{tabular}
\label{tab:mechanism_opt_settings}
\end{table}

For \cref{fig:lr_decay_test_acc}, the parameter settings differed based on the mechanisms explored. Constant LR schedules used the same settings as $\epsilon=18.9$ in \cref{tab:mechanism_opt_settings}. For the exploration of learning rate decay schedules, a server learning rate of $0.5$, client learning rate of $1.0$, and server momentum of $0.95$ were shared. The plot \cref{fig:lr_mom_eval} was generated from the same set of experiments.

\section{Background on Differential Privacy}
\label{sec:dpBackground}

In this paper we operate with the ``replace with zero'' variant of differential privacy~\citep[Defn. 2.1]{kairouz2021practical}, sated below for completeness purposes.

\begin{definition}[Differential privacy] 
Let $\calD$ be the domain of data records,  $\nul\not\in\calD$ be a special element, and let $\widehat{\calD}=\calD\cup\{\nul\}$ be the extended domain. 
A randomized algorithm $\calA:\widehat{\calD}^{n}\to\calS$ is $(\eps,\delta)$-differentially private if for any data set $D\in\widehat{\calD}^n$ and any neighbor  $D'\in\widehat{\calD}^n$ (formed from $D$ by replacing one record with $\nul$),
and for any event $S\in\calS$, we have 
\begin{align*}
    \Pr[\calA(D)\in S] &\leq e^{\eps} \cdot \Pr[\calA(D')\in S] +\delta,\ \ \text{and}\\
    \Pr[\calA(D')\in S] &\leq e^{\eps} \cdot \Pr[\calA(D)\in S] +\delta,
\end{align*}
where the probability is over the randomness of $\calA$.
\label{def:diiffP}
\end{definition}
In our algorithms, we would treat $\bot$ specially, and assume it corresponds to the all-zeros vector of appropriate dimensions. This definition extends naturally to other variants like Renyi differential privacy (RDP)~\cite{mironov2017renyi}, and zero Concentrated Differential Privacy (zCDP). For completeness purposes we provide the definition of zCDP we primarily use in the paper.

\begin{definition}[zero concentrated differential privacy]
Analogous to the definitiion of $(\epsilon,\delta)$-differential privacy in Definition~\ref{def:diiffP}, a randomized algorithm $\calA$ is $\rho$-zCDP if the condition on $\calA(D)$ and $\calA(D')$ in Definition~\ref{def:diiffP} are replaced with the following:
\begin{align*}
    \frac{1}{\alpha-1} \log \E_{s\sim \calA(D)}{\power{\frac{\Pr\left[\calA(D) = s\right]}{\Pr\left[\calA(D') = s\right]}}{\alpha}} &\leq \rho\alpha,\ \ \text{and}\\
     \frac{1}{\alpha-1} \log \E_{s\sim \calA(D')}{\power{\frac{\Pr\left[\calA(D') = s\right]}{\Pr\left[\calA(D) = s\right]}}{\alpha}} &\leq \rho\alpha.
\end{align*}
\label{def:zCDP}
\end{definition}

\end{document}